\documentclass[runningheads]{llncs}

\usepackage[mobile]{eccv}

\usepackage{eccvabbrv}
\usepackage{graphicx}
\usepackage[accsupp]{axessibility}  %
\usepackage{capt-of,xcolor,xspace,enumitem}
\usepackage{amsmath,amssymb,amsbsy,amsfonts,dsfont,pifont,bm,bbm,mathrsfs,mathtools,nicefrac}
\usepackage{algorithm,algpseudocode,listings}
\usepackage{booktabs,multirow,adjustbox,diagbox,threeparttable,tabularray}
\usepackage{makecell}

\usepackage{hyperref}

\usepackage{orcidlink}
\usepackage{wrapfig}
\usepackage[capitalize]{cleveref}  %
\crefname{section}{Sec.}{Secs.}
\Crefname{section}{Section}{Sections}
\crefname{table}{Tab.}{Tabs.}
\Crefname{table}{Table}{Tables}
\crefname{figure}{Fig.}{Figs.}
\Crefname{figure}{Figure}{Figures}
\crefname{equation}{Eq.}{Eqs.}
\Crefname{equation}{Equation}{Equations}
\newcommand{\bx}{\mathbf{x}}

\newcommand{\bI}{\mathbf{I}}

\newcommand{\bz}{\mathbf{z}}

\newcommand{\bff}{\mathbf{f}}
\newcommand{\bF}{\mathbf{F}}

\newcommand{\bc}{\mathbf{c}}

\newcommand{\bxi}{\boldsymbol{\xi}}

\newcommand{\bpsi}{\boldsymbol{\psi}}
\newcommand{\bzeta}{\boldsymbol{\zeta}}

\newcommand{\nR}{\mathbb{R}}

\newcommand{\nE}{\mathbb{E}}

\newcommand{\cL}{\mathcal{L}}

\newcommand{\cD}{\mathcal{D}}

\makeatletter
\DeclareRobustCommand\onedot{\futurelet\@let@token\@onedot}
\def\@onedot{\ifx\@let@token.\else.\null\fi\xspace}
\def\eg{e.g\onedot} 
\def\ie{i.e\onedot}

\def\Fig{Fig\onedot}   
\makeatother

\newcommand{\figref}[1]{\Fig~\ref{#1}}

\renewcommand{\eqref}[1]{Eq.~\ref{#1}}

\newcommand{\boldparagraph}[1]{\vspace{0.2cm}\noindent{\bf #1:} }

\newif\ifcomment
\commenttrue
\ifcomment
	\newcommand{\ag}[1]{ \noindent {\color{red} {\bf Andreas:} {#1}} }
	\newcommand{\yl}[1]{ \noindent {\color{cyan} {\bf Yiyi:} {#1}} }

\else
	\newcommand{\ag}[1]{}
	\newcommand{\yl}[1]{}
\fi

\newcolumntype{P}[1]{>{\centering\arraybackslash}m{#1}}
 
\newcommand{\method}{\texttt{TeFF}\xspace}

\newcommand{\tocite}[1]{{\color{red} [TOCITE]}}

\title{Learning 3D-Aware GANs from Unposed Images \texorpdfstring{\\}~with Template Feature Field}
\titlerunning{TeFF}

\author{
Xinya Chen$^{1}$,
~Hanlei Guo$^{1}$,
~Yanrui Bin$^{2}$,
~Shangzhan Zhang$^{1}$,
~Yuanbo Yang$^{1}$,
~Yue Wang$^{1}$,
~Yujun Shen$^{3}$,
~Yiyi Liao$^{1}$\thanks{Corresponding author.} 
}
\authorrunning{X.~Chen et al.}

\institute{$^{1}$Zhejiang University~$^{2}$The Hong Kong Polytechnic University~$^{3}$Ant Group}%

\begin{document}

{
\maketitle
\begin{center}
    \includegraphics[width=1.0\linewidth]{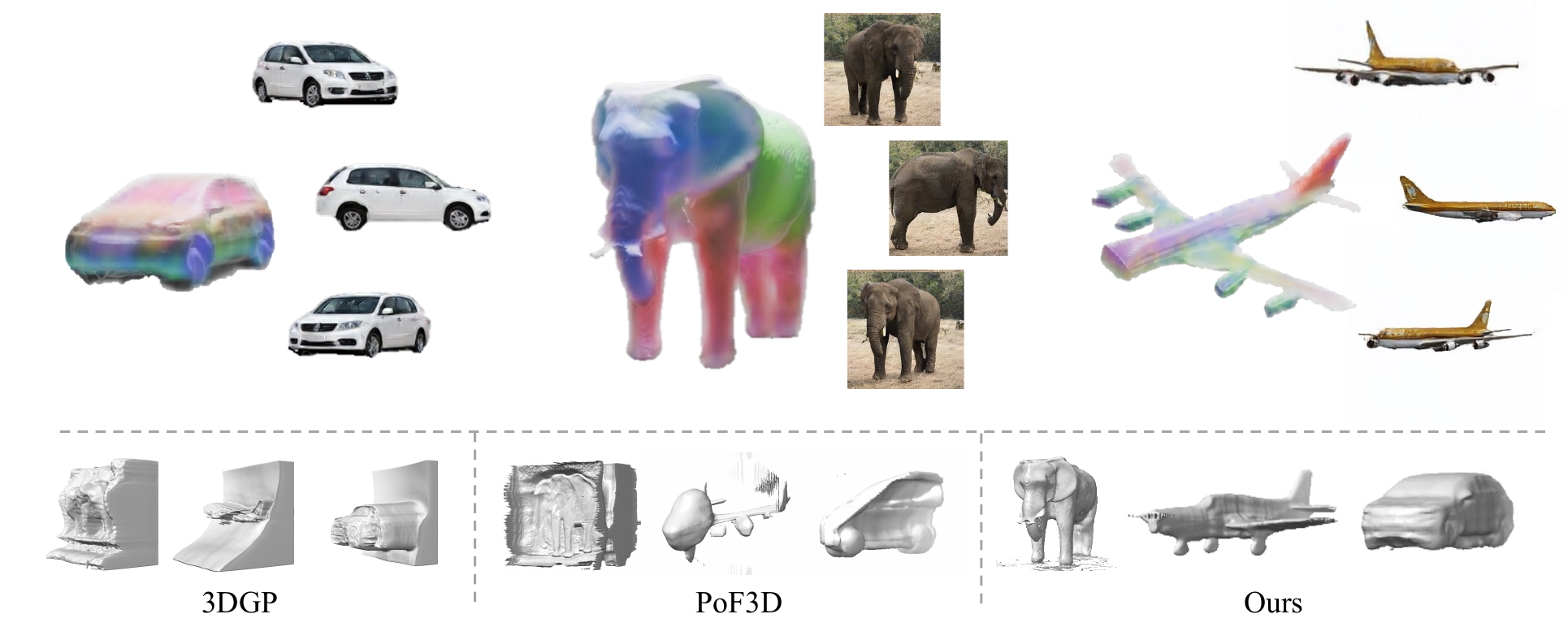}
    \captionsetup{type=figure}
    \caption{%
         We propose to perform on-the-fly pose estimation of training images with a learned template feature field. Our method enables learning complete 3D shapes from challenging datasets, including elephants, planes, and in-the-wild car datasets. While other methods have half or inaccurate geometry.
    }
    \label{fig:teaser}
\end{center}
}

\begin{abstract}

Collecting accurate camera poses of training images has been shown to well serve the learning of 3D-aware generative adversarial networks (GANs) yet can be quite expensive in practice.  
This work targets learning 3D-aware GANs from unposed images, for which we propose to perform on-the-fly pose estimation of training images with a learned template feature field (\method).
Concretely, in addition to a generative radiance field as in previous approaches, we ask the generator to also learn a field from 2D semantic features while sharing the density from the radiance field.  
Such a framework allows us to acquire a canonical 3D feature template leveraging the dataset mean discovered by the generative model, and further efficiently estimate the pose parameters on real data. 
Experimental results on various challenging datasets demonstrate the superiority of our approach over state-of-the-art alternatives from both the qualitative and the quantitative perspectives.
Project page: \urlstyle{same} \url{https://XDimlab.github.io/TeFF}.
 \keywords{3d-aware generation, semantic features, pose estimation}
\end{abstract}

\section{Introduction}\label{sec:intro}

3D assets are an important part of popular media formats such as video games, movies, and computer graphics. Compared to designing 3D content manually which is expensive and time-consuming, learning generative 3D models from data is a promising alternative to reduce the design effort. It is particularly attractive to learn 3D generative models from easy-to-obtained 2D images that generalize to different categories.

\begin{figure}[t]
     \def\mywidth{.33}
     \begin{tabular}{ccc}
     \includegraphics[width=\mywidth\linewidth]{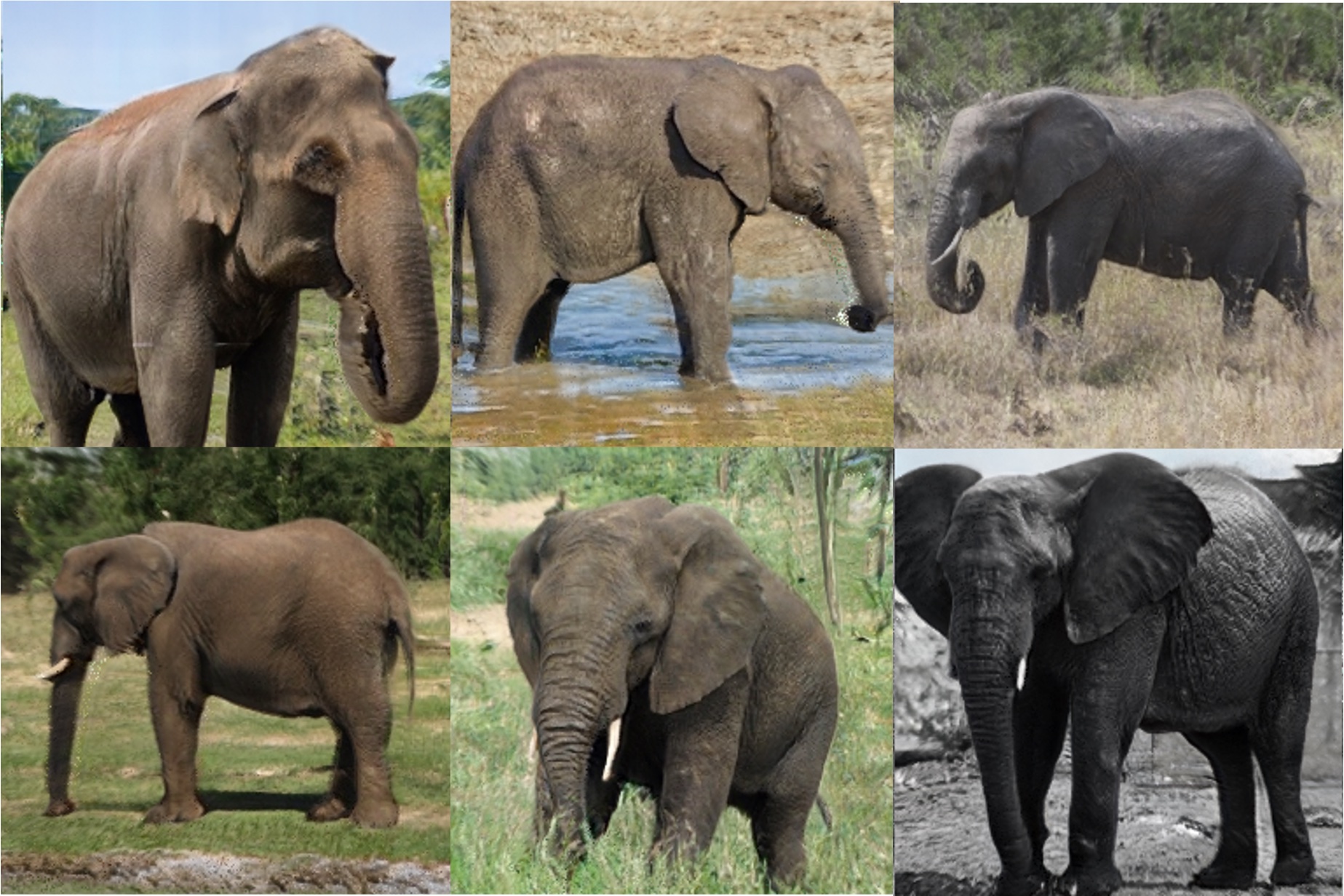} &
      \includegraphics[width=\mywidth\linewidth]{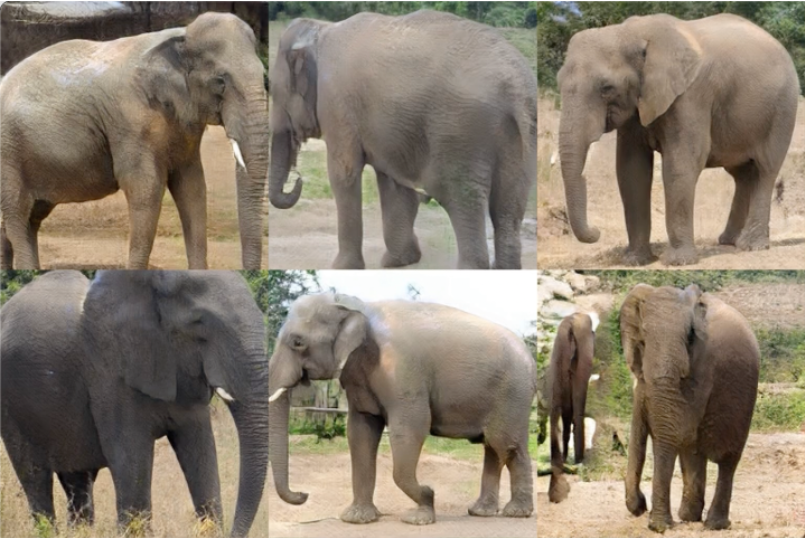} &
      \includegraphics[width=\mywidth\linewidth]{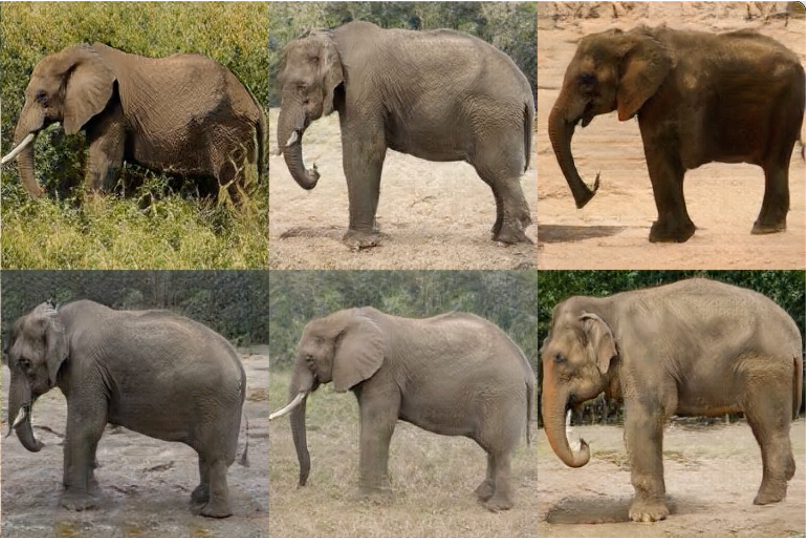} \\
      \begin{small}3DGP\end{small} &
      \begin{small}PoF3D\end{small} &
      \begin{small}Ours\end{small} 
     \end{tabular}
     \caption{\textbf{Qualitative Comparison} of 3DGP ($\{cols. 1-3\}$), PoF3D ($\{cols. 4-6\}$), Ours ($\{cols. 7-9\}$) when rendering in the same camera pose. Note that
objects generated by 3DGP and PoF3D face different directions with the same camera pose while ours face the same direction.}
   \label{fig:obj_pose}
    \end{figure}

Recent 3D-aware GANs have shown great success in this direction~\cite{Schwarz2020NeurIPS,Chan2020CVPR,Chan2021ARXIV}.
Most methods are designed for datasets with known camera pose distribution defined with respect to a canonical object pose. It has been proved that knowing the exact camera poses of real images further improves the learned geometry~\cite{EG3D,zhao2022gmpi}. 
However, estimating camera poses of real images requires domain-specific 3D knowledge about the object category,  which is difficult to acquire for most in-the-wild objects. 
To remove the assumption of known camera poses, there are some recent attempts to use a generator to jointly learn the camera pose distribution and the 3D contents. 
Despite achieving promising performance on faces and synthetic objects~\cite{Niemeyer2021THREEDV, shi2023pof3d}, these methods struggle in object categories with more complex pose distributions such as multi-peak distribution. This is mainly due to the fact that the generated camera poses and object poses are entangled in the 2D image space, leading to highly ill-posed solution space. For example, the objects may be generated facing different directions with a fixed camera so that the synthesized images can match the target distribution, as the result of 3DGP~\cite{3dgp} and PoF3D~\cite{shi2023pof3d} shown in \cref{fig:obj_pose}. This leads to incomplete geometry as shown in \cref{fig:teaser} as parts of the geometry are never observed.

In this work, we tackle this challenge by disentangling the estimation of camera pose distribution from the training of the 3D-aware GANs.
Our key idea is to learn a 3D semantic template feature field along with the generative model and define the pose estimation as an auxiliary task taking the template feature field as the canonical object space. This template feature field is automatically learned from self-supervised 2D features, \textit{e.g.}, DINO features\cite{caron2021emerging}, which are semantically aligned across instances of different appearances, thus enabling pose estimation. 
Specifically, We augment the generative radiance field with a semantic feature field while sharing the density. Then we acquire a template feature field leveraging the dataset mean discovered by the generative model. With the learned 3D template feature field and a 2D semantic feature of a real image, we estimate the camera pose of the real image by solving a 3D-2D pose estimation task. For robust and efficient pose estimation, we propose to project the 3D template to the 2D image space using a discretized set of camera poses and find the one with the best matching. This allows for achieving global optimum pose estimation when an ideal template is provided. To address the slow matching process when the camera pose discretization space is of high dimensional, we propose to discretize the azimuth and elevation only, and leverage phase correlation to efficiently solve scale and in-plane rotation.
We demonstrate that our method can generate full geometry even in complex pose distribution, including real-world cars, planes, and in-the-wild animals like elephants.

Our contributions can be summarized as follows, 
1) We present a novel 3D-aware generative model that jointly learns a semantic template feature field to enable pose estimation of real-world images on the fly, thus enabling training without known pose distribution.
2) We propose to efficiently solve the camera pose estimation by incorporating phase correlation for estimation scale and in-plane rotation.
3) Our model learns 3D-aware generative models on multiple challenging datasets, including real-world cars, planes, and elephants.

\section{Related Work}\label{sec:related}
\boldparagraph{3D-Aware Image Synthesis}
Recently, 3D-aware image synthesis has attracted growing attention by lifting the generator to the 3D space. A key to 3D-aware image synthesis is the choice of the underlying 3D representation. Early methods attempt to learn 3D voxel grid~\cite{Nguyen-Phuoc2019ICCV, Henzler2019ICCV} and mesh~\cite{Liao2020CVPR}, yet these discretized representations limit the image fidelity. More recent methods exploit neural radiance fields~\cite{Mildenhall2020ECCV} for 3D-aware image synthesis ~\cite{Schwarz2020NeurIPS,Chan2020CVPR,Chan2021ARXIV,Gu2021ARXIV,Jo2021ARXIV,Xu2021ARXIV,Zhou2021ARXIV,OrEl2021ARXIV,Xu2021NEURIPS,Pan2021NEURIPS,DENG2021ARXIV,Schwarz2022NEURIPS,bib:mimic3d,niemeyer2021giraffe, deng2022gram, xue2022giraffe, xiang2023gram, jo20233d, Chen_2023_ICCV, shin2023ballgan}. Albeit achieving photorealistic 3D-aware image synthesis, these methods focus on datasets with known camera pose distribution, \eg, human and cat faces or synthetic car datasets.

\boldparagraph{3D-Aware Image Synthesis from Unposed images}
There are a few attempts to address the task of 3D-aware unposed image synthesis~\cite{3dgp, Niemeyer2021THREEDV, shi2023pof3d, GET3D--}. 
CAMPARI\cite{Niemeyer2021THREEDV} designs the camera generator as a residual function to map a prior pose distribution to the target pose distribution. However, their performance is highly sensitive to the given pose prior. PoF3D~\cite{shi2023pof3d} frees the requirements of 3D pose priors by learning a pose-free generator which map a Gaussian distribution to the target distribution and learning a pose-aware discriminator in an adversarial manner. These methods only consider camera poses of two degrees of freedom (DoF), e.g., azimuth and elevation angles, based on the assumption that all cameras lie on a sphere. Recently, 3DGP   ~\cite{3dgp} extends the DoF from two to six by additionally modeling the field of view of the camera and the look-at point. \cite{GET3D--} model the translation and the rotation of the camera with 6 DoF. All these methods adopt a generator to jointly learn the camera pose distribution and the 3D contents. Despite performing well for data with a simple pose distribution, \eg, Gaussian distribution, these methods are prone to fail when the pose distribution becomes more complex. We instead learn a 3D semantic template feature field to solve the camera pose of each real image. Our method can generate full geometry even in complex pose distribution.

\boldparagraph{Neural Feature Fields}
Self-supervised models, like
DINO~\cite{caron2021emerging}, can solve video instance segmentation and tracking by calculating the similarity among features in adjacent frames.~\cite{amir2021deep} demonstrates that DINO features also work well on co-segmentation and point correspondence of different instances. Reconstruction methods have explored incorporating DINO features into neural fields for scene decomposition, editing, and semantic understanding~\cite{kobayashi2022distilledfeaturefields, lerf2023,yang2023emernerf,tschernezki22neural}. In this work, we explore the semantic-aware characteristics of such features to enable camera pose estimation across different instances within the same category.

\section{Method}\label{sec:method}
\begin{figure*}[t]
  \centering
  \includegraphics[width=\linewidth]{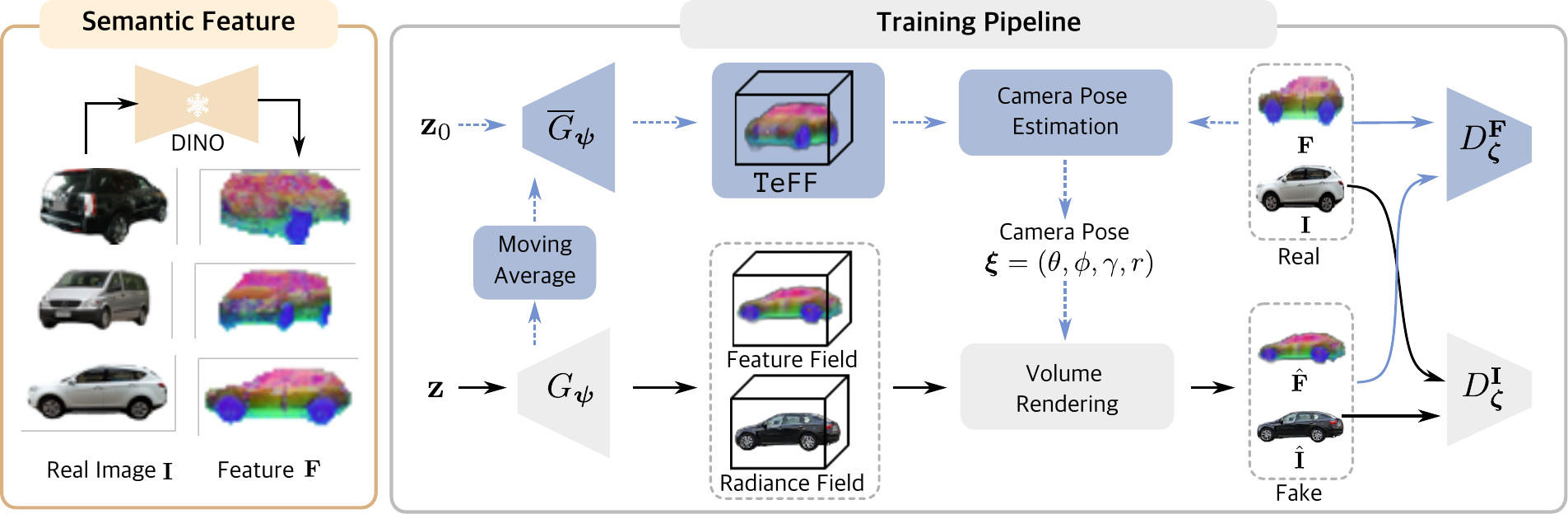}
   \caption{\textbf{Method Overview.} We augment the generative radiance field with a semantic feature field, enabling estimating camera poses of real images on the fly to facilitate the 3D-aware GAN training. Specifically, we map a randomly sampled noise vector to a radiance field and a semantic feature field. By taking the mean shape of the feature field, we obtain a 3D template feature field. This allows us to perform efficient 2D-3D pose estimation to estimate camera poses of real images, which are in turn fed into the generator to perform volume rendering. The blue part is the auxiliary task of pose estimation we introduced.
   }
   \label{fig:method_pipeline}
\end{figure*}

\figref{fig:method_pipeline} gives an overview of our method. 
We augment a 3D-aware GAN using our Template Feature Fields (\method) to enable learning from real-world images without camera pose distributions.
The key to our approach is the joint learning of the generative radiance field and a semantic feature field, where a 3D template feature field can be extracted to solve the camera poses of the real images on the fly. In the following sections, we first introduce the offline semantic feature extraction in \cref{sec:feature_extraction}. Next, we present our generator augmented with \method in \cref{sec:teff}. We further elaborate on the camera pose estimation of the real images based on the 2D feature maps and the 3D template feature field in \cref{sec:pose}, followed by the discriminator and implementation details in \cref{sec:training} and \cref{sec:detail}.

\subsection{Semantic Feature Extraction}
\label{sec:feature_extraction}

In this paper, we refer to semantic features as feature maps that are semantically aligned across instances of different appearances and poses, as illustrated in \cref{fig:method_pipeline} (left). For example, the car wheels share the same features across different instances. Existing methods have demonstrated that such features can be leveraged to establish correspondences between different instances~\cite{amir2021deep,caron2021emerging}, potentially enabling their relative pose estimation. In this work, we propose to extract semantic features from each real image to enable its camera pose estimation.

Specifically, we extract semantic features $\bF\in\nR^{H\times W \times F}$ from real RGB images $\bI\in\nR^{H\times W \times 3}$ based on DINO\cite{caron2021emerging}, as we observe that DINO generalizes well and provides meaningful semantic features for various categories. In practice, we feed RGB images to the DINO model to get DINO feature maps. Then we extract foreground masks from these RGB images and apply the mask to DINO feature maps. This allows us to focus on the features of foreground instances.

\subsection{Generator with \method}
\label{sec:teff}

\boldparagraph{Generative Radiance and Feature Fields}
We build our generator based on EG3D \cite{EG3D} but with a key difference in that the model jointly generates radiance fields and feature fields. Conditioned on a noise $\bz\in\nR^{M}$ sampled from a Gaussian distribution, the generator $G_{\bpsi}$ generates a radiance field and a feature field that maps a 3D point $\bx$ to a RGB value $\bc$, feature value $\bff$, and density value $\sigma$:

\begin{equation}
G_{\bpsi}: \nR^3\times\nR^{M} \rightarrow \nR^3 \times \nR^F \times\nR^+ \quad (\bx, \bz) \mapsto (\bc, \bff, \sigma)
\end{equation} 
where $F$ denotes the dimension of the feature vector.
In practice, the generator produces two tri-planes, one for the color and density, the other for the feature.

The generative radiance field is rendered to a 2D image via the volume rendering operation. Given an estimated camera pose $\bxi$ (discussed in \cref{sec:pose}), the generator is queried on $H\times W$ rays with $N$ sampling points on each ray. Let $\{\bc_i, \bff_i, \sigma_i\}_{i=1,\dots,N}$ denote the queried color, feature, and density values of one ray, the pixel color $\bc_r$ and feature $\bff_r$ at the corresponding ray is obtained via: 
\begin{equation}
    \begin{gathered}
 \pi: (\nR^{3}\times\nR^+)^N \rightarrow \nR^3 \qquad \{(\bc_i,\bff_i,\sigma_i)\} \mapsto \bc_r,\bff_r \nonumber \\
\bc_r = \sum_{i=1}^N \, T_i \, \alpha_i \, \bc_i \quad
\bff_r = \sum_{i=1}^N \, T_i \, \alpha_i \, \bff_i \quad
T_i = \prod_{j=1}^{i-1}\left(1-\alpha_j\right) \quad
\alpha_i = 1-\exp(-\sigma_i \delta_i) \nonumber
    \label{eq:volume_rendering}
    \end{gathered}
\end{equation} 
where $T_i$ and $\alpha_i$ denote the transmittance and alpha value at a sample point $\bx_i$, $\delta_i$ denotes the distance interval between two adjacent sample points. Note that the same density values are shared for rendering the color $\bc_r$ and the feature $\bff_r$. By compositing all ray values we obtain a color image $\hat{\bI}\in\nR^{H\times W \times 3}$ and a feature map $\hat{\bF}\in\nR^{H\times W \times F}$.

Following EG3D~\cite{EG3D}, we apply a super-resolution module to the rendered RGB image for upsampling via neural rendering. Note that the super-sampling module is not shown in \cref{fig:method_pipeline} for brevity. We do not perform super-resolution modules to the rendered feature maps as we observe that low-resolution feature maps are sufficient for our purpose of pose estimation.

\boldparagraph{Template Feature Field} 
We propose to leverage the mean feature fields automatically discovered by the generator to enable the subsequent camera pose estimation. This mean feature field can be considered as a template for the full category.
To obtain this template feature field, we first apply the exponential moving average to the generator $G_{\bpsi}$ during training to get $\overline{G}_{\bpsi}$. Next, we feed the mean noise vector $\bz_\textbf{0}$ to the moving averaged generator $\overline{G}_{\bpsi}$ to decode the feature and density for the template. 

\boldparagraph{Background Generator}
In order to estimate camera poses focusing on the objects, we disentangle the foreground and background generation.
Following VoxGRAF~\cite{Schwarz2022NEURIPS}, we generate the background of the RGB image with a 2D GAN for efficiency. Specifically, we use the StyleGAN2 generator\cite{Karras2019stylegan2} with reduced channel size as modeling the background requires less capacity than generating the full image.
We use the same latent code $\bz$ of the foreground for the background generator to allow for modeling correlations like lighting between the two. 
The final RGB image is obtained by alpha compositing the rendered foreground image and the 2D background.

\begin{figure*}[t]
  \centering
  \includegraphics[width=\linewidth]{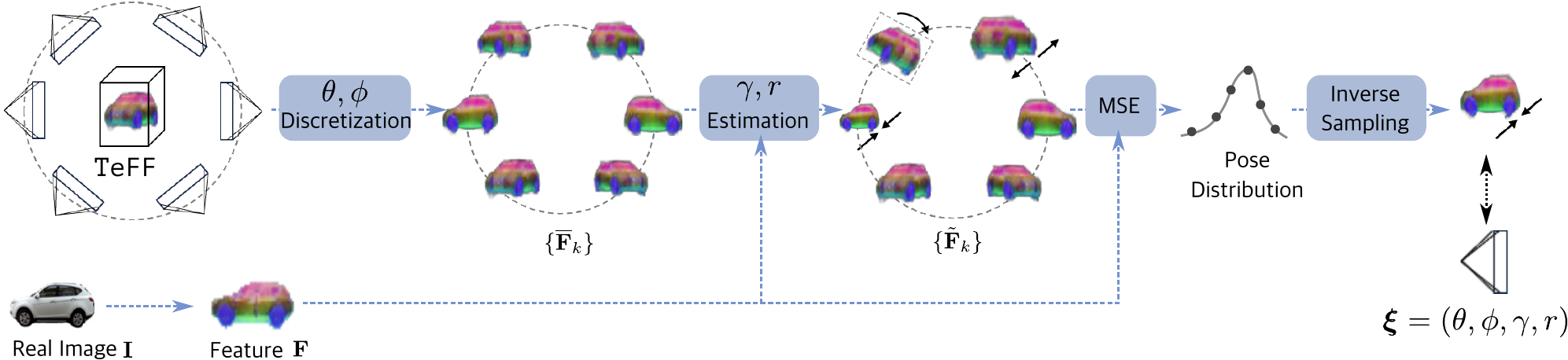}
   \caption{\textbf{Camera Pose Estimation.} We leverage the template feature field to estimate camera poses of 2D real images. We discretize the azimuth $\theta$ and elevation $\phi$ angles and render the feature field from these discretized camera poses. Then we use phase correlation to estimate the scale and the in-plane rotation in the 2D image space and warp each of the rendered templates based on the solution. We calculate the mean square error between the warped rendering and the real feature and further obtain the probability distribution function of the camera pose. Finally, we sample the camera pose using inverse sampling.}
   \label{fig:pipeline}
\end{figure*}

\subsection{Camera Pose Estimation}
\label{sec:pose}

We leverage the 3D template feature field to estimate camera poses of 2D real images, which can be formulated as a 2D-3D camera pose estimation task. This is possible since the features are semantically aligned across different instances.
The task of 2D-3D matching is non-trivial. Many existing 2D-3D pose matching methods require establishing 2D-3D correspondences~\cite{Peng2018PVNetPV}, followed by solving a Perspective-n-Point (PnP) problem for pose estimation. However, in our preliminary trials, it is non-trivial to establish correct 2D-3D correspondences (e.g., the left leg of an elephant in the 2D image might be matched to a right leg in the template), thus the estimated poses are unreliable. To tackle this challenge, we propose an efficient yet accurate way of solving the camera pose via grid search. The core idea is to discretize the camera pose space and select the one that renders a feature map that best matches the real feature. We now introduce how we define the camera model and how the grid search is efficiently implemented.

\begin{wrapfigure}[13]{r}{0.35\textwidth}
  \centering
  \includegraphics[width=0.35\textwidth]{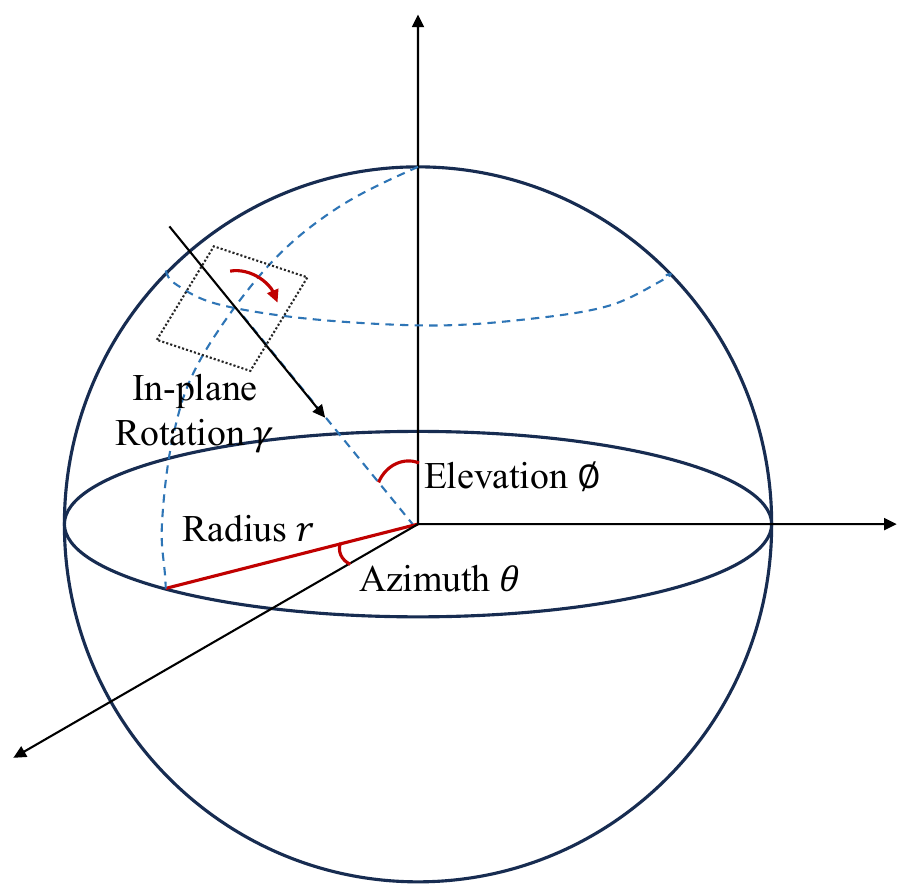}
  \caption{\textbf{Camera Model.} }
   
   \label{fig:failure}
\end{wrapfigure}
\boldparagraph{Camera Model}
We assume our camera lies on a sphere and looks at the sphere center, with freedom degrees of azimuth, elevation, in-plane rotation, and sphere radius. More formally, we parameterize the camera pose $\bxi = (\theta, \phi, \gamma, r)$. Here, $r$ denotes the radius of the sphere. $\theta\in[0,2\pi]$ and $\phi\in[0,\pi]$ denote the azimuth and elevation angles of the camera's translation in the spherical coordinate system. By looking at the sphere center from the direction determined by $\theta$ and $\phi$ and performing the in-plane rotation $\gamma$, we can determine the camera's rotation. Note that, $r$ determines the scale of objects in the image, we use scale to refer it in the following. 

\boldparagraph{Azimuth and Elevation Discretization} 
With this 4 DoF camera model, 
a na\"ive grid search can be implemented by discretizing all four variables $(\theta, \phi, \gamma, r)$ to form a large number of camera poses, and render the feature field into a set of feature maps, followed by comparing them with a feature map of the real image and selecting the best-matching one. This exhaustive grid search allows for finding the best-matching pose among the candidates, yet it is extremely computationally expensive. Hence, we propose to combine grid search with the efficient phase correlation approach~\cite{Kuglin1975ThePC}, allowing for efficiently estimating the camera pose with a high accuracy.
Specifically, we discretize the azimuth $\theta$ and elevation $\phi$ angles into $N_\theta$ and $N_\phi$ values, respectively. Next, we render the template feature fields from these discretized azimuth $\theta$ and elevation $\phi$ angles, yielding a set of 2D template features $\{\overline{\bF}_k\}_{k=1}^{N_\theta \times N_\phi}$. 

\boldparagraph{Scale and In-plane Rotation Estimation} Given a semantic feature map $\bF$ of a real image $\bI$, we may retrieve its best-matching template feature from 
$\{\overline{\bF}_k\}$ and assign the corresponding pose to $\bI$. However, this does not model scale $r$ and in-plane rotation $\gamma$. Therefore, we further estimate the $r$ and $\gamma$ between the real feature map $\bF$ and \textit{each} 2D template feature $\overline{\bF}_k$. Considering the efficiency, we employ phase correlation (PC)~\cite{Kuglin1975ThePC}, a classical frequency-domain-based approach that allows for estimating the scale and in-plane rotation between two images efficiently, without relying on any explicit correspondence or initialization. This is because PC estimates scale and in-plane rotation following the same idea as grid search, yet the calculations are performed in the frequency domain for fast computation. 
Based on the estimated scale $r$ and in-plane rotation $\gamma$, we warp each 2D template image $\overline{\bF}_k$ to get $\tilde{\bF}_k$. Note that, the idea of performing image registration in the frequency domain is also studied in other works for different purposes \cite{zhao2022tex, chen2020deep}.

\boldparagraph{Camera Pose Sampling} We calculate the mean square error (mse) between each $\tilde{\bF}_k$ and the real semantic feature map $\bF$ as: 
\begin{equation}
e_k(\tilde{\bF}_k, \bF) =  \|\tilde{\bF}_k - \bF\|_2^2
\end{equation}
Ideally, we can assign the camera pose of $\tilde{\bF}_k$ corresponding to the lowest $e_k$ to the real feature map $\bF$. However, the feature template may not exactly match each real image, especially at the early training stage, occasionally leading to inaccurate pose estimation. Therefore, we form a probability distribution function (PDF) over all possible $N_\theta \times N_\phi$ poses and sample from it during training.
Specifically,
we sample a camera pose using inverse sampling according to the PDF of the camera pose. The pose probability $p(k)$ is calculated by:
\begin{equation}
    p(k) = softmax(-{e_k(\tilde{\bF}_k, \bF)}*\tau)
\end{equation}
where $\tau$ is the temperature, which controls the sharpness of the probability distribution function. At the early of training, we use low temperature $\tau$ and then increase the temperature linearly. 
Since we discretize the camera pose space,  we further add a slight Gaussian noise to the sampled pose. This allows us to sample continuous camera pose locations from each discretized bin.
During inference, we select the camera pose with the maximum likelihood.

\subsection{Discriminator and GAN Training}
\label{sec:training}

\boldparagraph{Discriminator}
In order to train our feature field, we add a feature discriminator $D_{\bzeta}^\bF$ on top of the original image discriminator $D_{\bzeta}^\bI$. $D_\psi^\bI$ is used to discriminate real RGB images and fake RGB images, whereas $D_{\bzeta}^\bF$ learns to discriminate the semantic features. Following~\cite{EG3D}, we adopt the dual discriminator with camera pose condition for $D_{\bzeta}^\bI$, i.e., the input to $D_{\bzeta}^\bI$ is an upsampled low-resolution RGB image, a high-resolution RGB image, and the corresponding camera pose. The input to $D_{\bzeta}^\bF$ is the low-resolution RGB image and the corresponding semantic feature map, to encourage the learned feature field to match with the corresponding radiance field. We stop the gradient backpropagation from $D_{\bzeta}^\bF$ into the RGB values following \cite{sun2022fenerf}.

\noindent\textbf{GAN Training} Given the latent code $\bz$, a real image $\bI$ and its corresponding semantic feature $\bF$ sampled from the real-data distribution $p_\cD$, we train our model using non-saturated GAN loss with R1 regularization~\cite{Mescheder2018ICML}.
\begin{align}
\cL  = &
\nE_{\bz\sim \mathcal{N}(0, \mathbf{1})}
\left[f(D_{\bzeta}^\bI\left(G_{\bpsi}(\bz)\right))
\,+\, f(D_{\bzeta}^\bF\left(G_{\bpsi}(\bz)\right))
\right]\nonumber \\
&\,+\, \nE_{\bI,\bF \sim p_{\cD}}
\left[
f(-D_{\bzeta}^\bI(\bI))
\,+\, \lambda {\Vert \nabla D_{\bzeta}^\bI(\bI)\Vert}^2
\,+\, f(-D_{\bzeta}^\bF(\bF))
\,+\, \lambda {\Vert \nabla D_{\bzeta}^\bF(\bF)\Vert}^2
\right] \nonumber 
\end{align}

\boldparagraph{Training Strategy} The pose estimation is done on the fly in the early stage. We observe that once the model learns a reasonable template, we can use the camera poses estimated from a fixed template for the remaining training. We update the template once every 16 iterations before 3k iterations and then update the template once epoch.

\section{Experiments}\label{sec:exp}
\boldparagraph{Datasets} We evaluate on four datasets, including Shapenet Cars~\cite{Chang2015ARXIV,EG3D}, CompCars~\cite{CompCars2015CVPR}, SDIP Elephant~\cite{mokady2022selfdistilled}, and LSUN Plane~\cite{Yu2015ARXIV}. Shapenet Cars is a synthetic dataset that contains car images with ground truth camera poses. Note that the ground truth poses are only used for evaluation for all methods. As the camera poses of the original dataset are roughly uniformly distributed, we consider a more challenging scenario by building a sub-dataset where the pose distribution has multiple peaks, with around 103K images. CompCars contains 136k unposed images capturing the entire cars with different styles. We mask the background to black and filter the data with bad mask estimation and extreme scale, leading to 110k images. We use the SDIP Elephant dataset following~\cite{3dgp} and filter the elephants with the half body, which contains around 20k unposed images. LSUN Plane is a dataset that contains unposed images of different planes. We mask the background to white and filter the data with extreme scale and occluded planes, leading to 130k images. In the experiments, we use the resolution of 256$\times$256 for SDIP Elephant, CompCars and LSUN Plane, and 128$\times$128 for Shapenet Cars.  

\boldparagraph{Baselines} We consider three baselines: EG3D~\cite{EG3D}, PoF3D~\cite{shi2023pof3d} and 3DGP~\cite{3dgp}.
EG3D is a state-of-the-art 3D-aware GAN for posed images, whereas PoF3D and 3DGP both learn from unposed images while jointly learning the camera pose distribution.
For EG3D we uniformly sample pose with azimuth angle sampled from 360 degrees and elevation from 85 degrees to 95 degrees. Note that the original EG3D conditions the generator and discriminator on the camera poses. We do not use this pose condition for EG3D due to the lack of image-pose pair data. For 3DGP, we set the prior azimuth angle with a uniform distribution of 360 degrees and set the prior elevation angle with a uniform distribution from 85 degrees to 95 degrees.

\boldparagraph{Metrics} We use three metrics to evaluate the performance, including Frechet Inception Distance (FID)~\cite{Heusel2017NIPS}, Depth Error~\cite{EG3D}, and Kullback-Leibler (KL) Divergence. FID is measured between 50K generated images and all real images. 
Note that, we evaluate the FID using two sets of camera poses. As the camera pose distributions estimated by different methods are different,
we first sample images using the ground truth pose distribution for evaluation as ``FID$_{gt}$''. This ensures a fair comparison of different methods.
In cases where the ground truth pose distribution is not available on CompCars, SDIP Elephant and LSUN Plane, we render fake images from a uniform distribution with azimuth angle sampled from 360 degrees and evaluate their fidelity as ``FID$_{360}$''. This is rationale as those datasets contains images captured from 360 degree azimuth angles. We further evaluate FID using the estimated camera poses, denoted as ``FID$_{est}$'' to reflect the image fidelity at the learned camera distributions.
Depth Error is used to assess the quality of geometry. Following~\cite{EG3D}, we use a monocular depth estimation algorithm~\cite{Wei2021CVPR} to get the pseudo depth and use a mask estimation algorithm~\cite{Wu2019GH} to obtain the foreground mask. We randomly generate 10k images and only evaluate the depth error of the foreground objects. 
We normalize the depth by subtracting the mean depth of the current sample and dividing the standard deviation depth of the dataset.
Similar to FID we also evaluate ``Depth$_{gt}$''/``Depth$_{360}$'' and ``Depth$_{est}$''.
The standard deviation depth uses the one calculated in the estimated camera pose setting since the depth estimation and mask estimation may fail in the 360 camera pose setting.
For pose distribution estimation, we randomly sample 10k ground truth poses and estimated poses, and then calculate their KL Divergence. We adopt this distribution loss since the baseline methods do not provide an estimated camera pose for each real image but learn a distribution.

\subsection{Implementation Details}
\label{sec:detail}

In our implementation, We extract the semantic features from DINO\cite{caron2021emerging} using real images of 256$\times$256 resolution and get the DINO features of 64$\times$64 resolution. We apply Principal Component Analysis to the feature maps and keep the first three principal components of the original DINO features for training efficiency. This means the feature channel $F$ is set to $3$. 

The standard deviation of the noise is 1/6 of the discrete interval. 
We discretize the azimuth $\theta$ and elevation $\phi$ angles into $N_\theta$ and $N_\phi$ values, respectively. The azimuth range is 360 degrees and $N_\theta$ is 36 for all datasets. The elevation range is 180 degrees and $N_\phi$ is 18 for the ShapeNet Cars dataset. For other datasets, the elevation range is from 85 to 95 degrees and $N_\phi$ is 3 since the elevation variation is small.

\subsection{Comparison to the Baselines}
\begin{figure}[t!]
     \def\mywidth{11cm}
     \begin{tabular}{P{0.5cm}P{\mywidth}}
     \rotatebox{90}{EG3D} &\includegraphics[width=\mywidth]{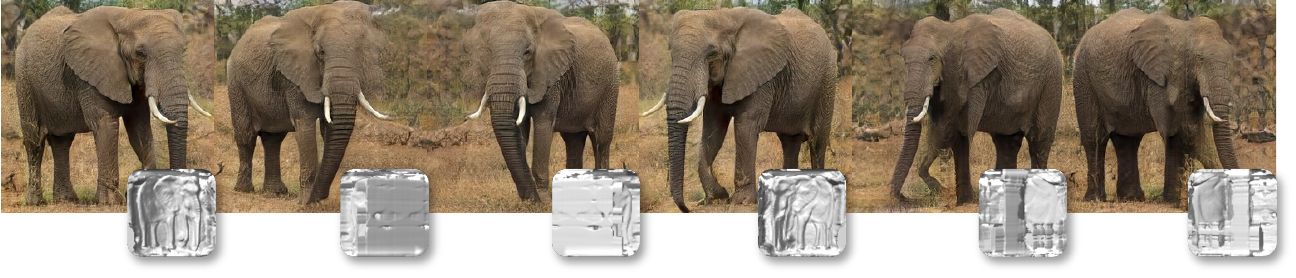} \\
    \rotatebox{90}{3DGP}
    &\includegraphics[width=\mywidth]{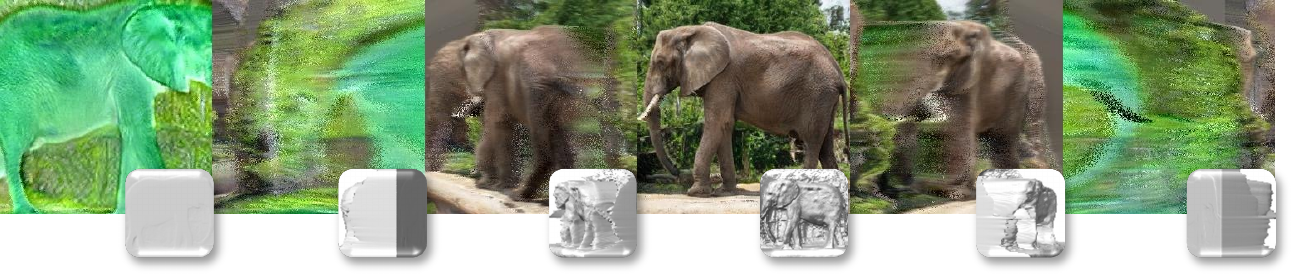} \\
    \rotatebox{90}{PoF3D}
    &\includegraphics[width=\mywidth]{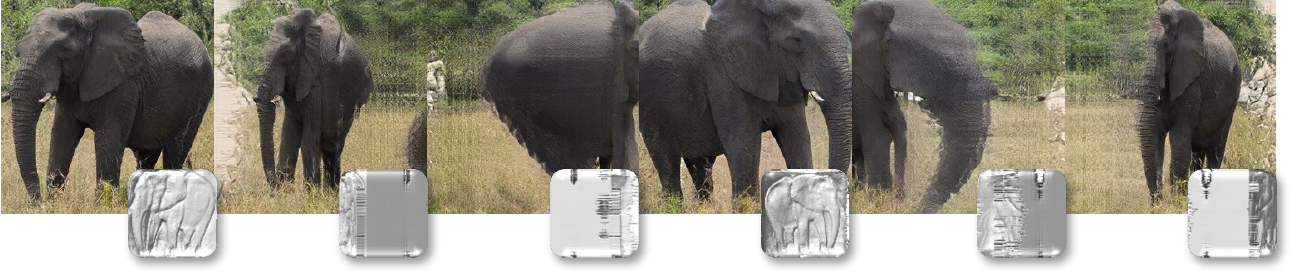} \\
     \rotatebox{90}{Ours} &\includegraphics[width=\mywidth]{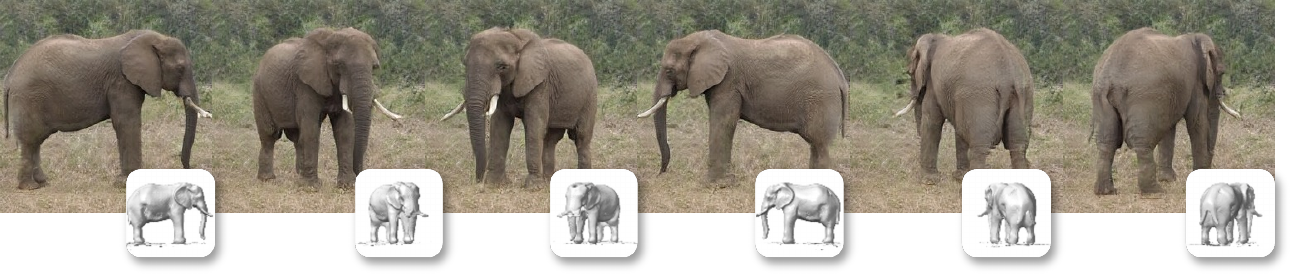}\\
     \end{tabular}
     \caption{\textbf{Qualitative Comparison on SDIP Elephant.} We show each sample from 360 viewing directions.}
   \label{fig:comparison_elephant}
    \end{figure}

\begin{figure}[h]
     \def\mywidth{11cm}
     \begin{tabular}{P{0.5cm}P{\mywidth}}
     \rotatebox{90}{EG3D} &\includegraphics[width=\mywidth]{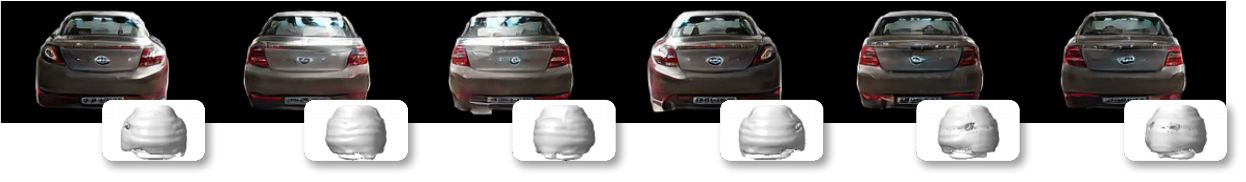} \\
    \rotatebox{90}{3DGP}
    &\includegraphics[width=\mywidth]{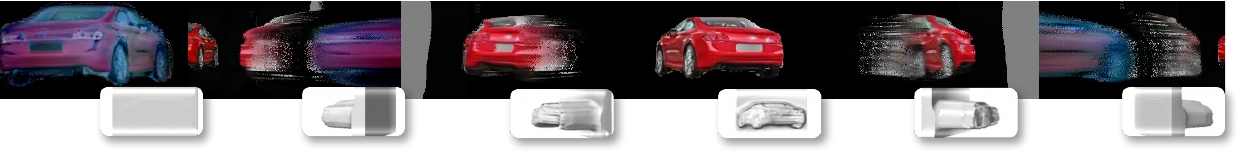} \\
    \rotatebox{90}{PoF3D}
    &\includegraphics[width=\mywidth]{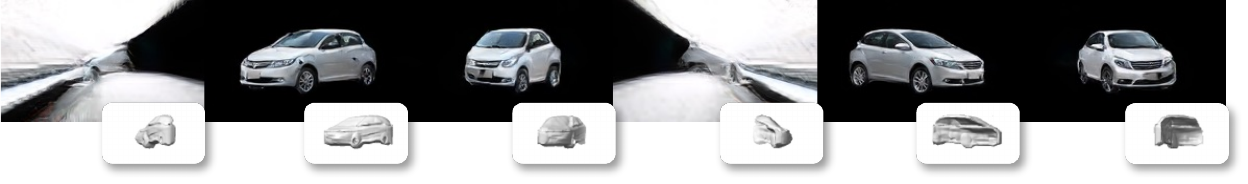} \\
     \rotatebox{90}{Ours} &\includegraphics[width=\mywidth]{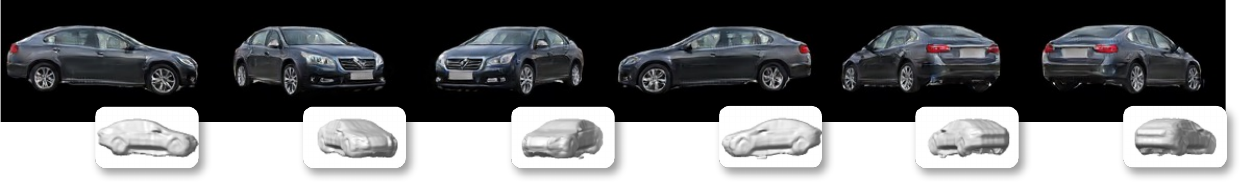}\\
     \end{tabular}
     \caption{\textbf{Qualitative Comparison on CompCars.} We show each sample from 360 viewing directions.}
   \label{fig:comparison_CompCars}
    \end{figure}

\begin{figure}[h]
     \def\mywidth{11cm}
     \begin{tabular}{P{0.5cm}P{\mywidth}}
     \rotatebox{90}{EG3D} &\includegraphics[width=\mywidth]{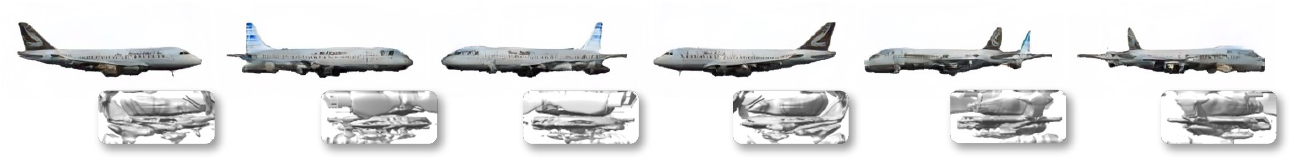} \\
    \rotatebox{90}{3DGP}
    &\includegraphics[width=\mywidth]{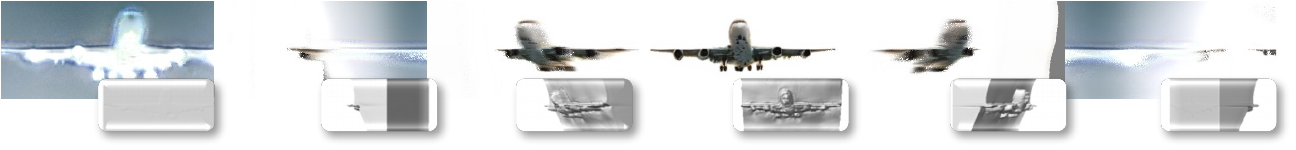} \\
    \rotatebox{90}{PoF3D}
    &\includegraphics[width=\mywidth]{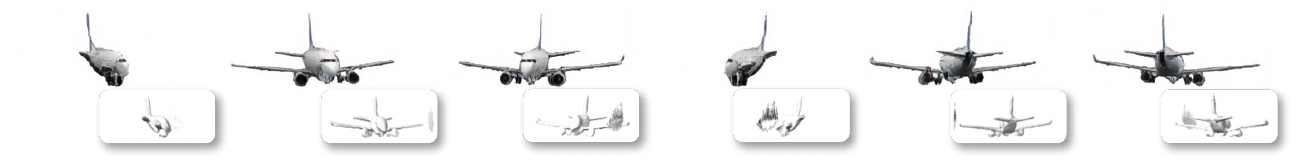} \\
     \rotatebox{90}{Ours} &\includegraphics[width=\mywidth]{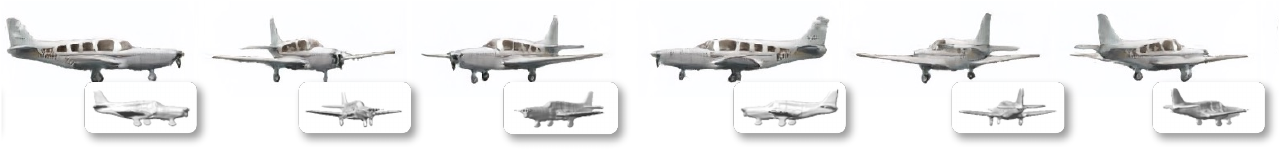}\\
     \end{tabular}
     \caption{\textbf{Qualitative Comparison on LSUN Plane.} We show each sample from 360 viewing directions with a 60-degree interval.} 
   \label{fig:comparison_plane}
    \end{figure}

\boldparagraph{Qualitative Comparison} \figref{fig:comparison_elephant}, \figref{fig:comparison_CompCars}, \figref{fig:comparison_plane} shows the rendering comparison against the baselines on SDIP Elephant, CompCars, and LSUN Plane. We render the image from 360 degrees at 60-degree intervals. The bottom right of the image illustrates the corresponding geometry. EG3D~\cite{EG3D}, 3DGP~\cite{3dgp}, and PoF3D~\cite{shi2023pof3d} perform well in some specific views but fail to perform 360-degree rendering with consistency due to the incomplete or inaccurate geometry. Instead, our method can perform 360-degree rendering with good fidelity. Besides, our method allows for learning the complete geometry with high quality, whereas the baselines fail to reconstruct the geometry correctly or only reconstruct half of the object. 

\boldparagraph{Quantitative Comparison} We report the quantitative evaluation of baselines and our method in \cref{tab:quant_car} and \cref{tab:quant_ele}. The result suggests that EG3D trained with unknown camera pose distribution can achieve a low FID, yet results in the worst geometry. Note that EG3D achieves the best FID$_{360}$ since the azimuth angle is uniformly sampled from 360 degrees during training. Despite having high fidelity for all surrounding viewpoints, EG3D does not recover the correct geometry and thus yields bad Depth$_{360}$, e.g., as shown in \cref{fig:comparison_elephant}. 3DGP and PoF3D achieve the best FID when evaluated on their estimated camera poses, whereas the performance drops significantly when evaluated using camera poses uniformly sampled on CompCars, SDIP Elephant and LSUN Plane. The reason is that 3DGP and PoF3D learn a collapsed camera pose distribution, \ie, it only covers a small range of the underlying true pose distribution. Our method instead retains a similar performance when evaluated using the uniform distribution, i.e., our FID$_{360}$ and FID$_{est}$ are close, thanks to the fact that our method learns to recover a full object in the 3D space for reasonable rendering from all surrounding viewpoints. This is also verified by the fact that our method performs the best in terms of the Depth$_{360}$. Note that 3DGP achieves the best Depth$_{est}$ since it uses the depth maps as supervision during training, and we use the same monocular depth estimation algorithm to provide depth supervision for 3DGP and for evaluating its depth performance. Despite not relying on any depth supervision, our method even achieves comparable Depth$_{est}$ compared to 3DGP on the car datasets in \cref{tab:quant_car}. 

\small
\begin{table}[t]
  \centering
  \resizebox{\textwidth}{!}{
  \begin{tabular}{c|cccc|cccc}
    \hline
     & \multicolumn{4}{c|}{Multiple-Peak Shapenet Cars}& \multicolumn{4}{c}{CompCars} \\
    Method & Depth$_{gt}$$\downarrow$ & Depth$_{est}$ $\downarrow$ & FID$_{gt}$ $\downarrow$ &  FID$_{est}$ $\downarrow$
    & Depth$_{360}$$\downarrow$ & Depth$_{est}$ $\downarrow$ & FID$_{360}$$\downarrow$ &  FID$_{est}$ $\downarrow$\\
    \hline
    EG3D  & 0.61 & 0.61 & 7.25 & 11.89 & 0.95 & 0.95 & \textbf{7.06} & 7.06\\
    3DGP &4.84 & \textbf{0.46} & 139.48 & \textbf{4.93} & 4.02 & 0.37 & 187.20  &  8.34  \\
    PoF3D & 0.65 & 0.70 & 12.72 & 6.45 & 10.31 & 0.70 & 44.52 & \textbf{6.62}  \\ 
    Ours  & \textbf{0.53} & 0.52 & \textbf{5.95} & 6.55 & \textbf{0.31} & \textbf{0.29} & 27.71 & 20.60  \\ %
    \hline
  \end{tabular}
   }
  \caption{%
    \textbf{Quantitative Comparison} on Shapenet Cars and CompCars.
  }
  \label{tab:quant_car}
\end{table}

\begin{table}[t]
  \centering
  \resizebox{\textwidth}{!}{
  \begin{tabular}{c|cccc|cccc}
    \hline
     & \multicolumn{4}{c|}{SDIP Elephant} & \multicolumn{4}{c}{LSUN Plane}\\
    Method & Depth$_{360}$$\downarrow$ & Depth$_{est}$ $\downarrow$ & FID$_{360}$$\downarrow$ &  FID$_{est}$ $\downarrow$
    & Depth$_{360}$~$\downarrow$ & Depth$_{est}$~$\downarrow$ & FID$_{360}$   $\downarrow$ &  FID$_{est}$ $\downarrow$\\
    \hline
    EG3D  & 1.10 & 1.10 & 6.03 & 6.03 & 1.19 & 1.19 & \textbf{7.27} & 7.27\\
    3DGP & 3.29 & \textbf{0.30} & 196.04 & \textbf{2.18} & 3.84 & \textbf{0.33} & 196.29 & \textbf{3.79}  \\
    PoF3D & 3.14 & 1.08 & 36.32 & 3.04 & 1.37 & 1.07 & 16.26 & 6.07 \\
    Ours  & \textbf{0.60}& 0.68 & \textbf{5.51} & 5.44 & \textbf{0.78} & 0.78 & 14.20 & 16.55 \\
    \hline
  \end{tabular}
   }
  \caption{%
    \textbf{Quantitative Comparison} on SDIP Elephant and LSUN Plane.
  }
  \label{tab:quant_ele}
\end{table}

\begin{wraptable}[10]{r}{0.35\textwidth}
    \centering
    \begin{tabular}{c|ccc}
      \hline
      Method & $\theta$ KL~$\downarrow$ & $\phi$ KL  ~$\downarrow$\\

      3DGP & 40.4571 &  39.3625  \\
      PoF3D & 4.4829 & 0.5495   \\
      Ours & \textbf{0.0555} & \textbf{0.0696} \\
      \hline
    \end{tabular}
    \caption{\textbf{Pose Distribution Comparison.}}
    \label{tab:pose_comp}
\end{wraptable}

\begin{figure}[t!]
     \def\mywidth{12cm}

     \includegraphics[width=\mywidth]{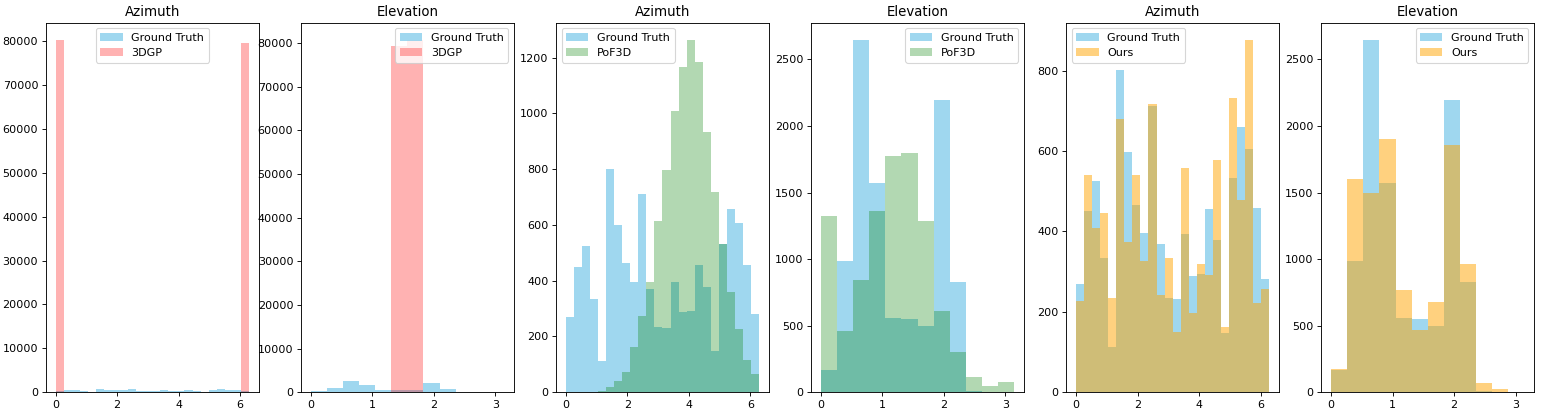} \\
     \caption{\textbf{Pose Distribution Comparison on Shapenet Cars.} }
   \label{fig:pose_comparison}
    \end{figure}

\boldparagraph{Pose Distribution Comparison} 
We use the ShapeNet Cars with ground truth poses to evaluate the accuracy of the estimated camera pose distribution.
We discretize the azimuth angle and the elevation angle into 24 and 12 intervals, and calculate the discrete probability distributions over these intervals. \figref{fig:pose_comparison} demonstrate the comparison between the GT and estimated pose distribution of different methods regarding azimuth and elevation angles. 
It is shown that the pose distribution of 3DGP is collapsed to a small range. PoF3D covers a larger range but fails to capture the multi-modal distribution: the learned distribution remains Gaussian, similar to its initialization. In contrast, our estimated pose distribution is notably more accurate. 
This is also verified by the quantitative comparison of the KL divergence in \cref{tab:pose_comp}, where our distribution is significantly better than that of 3DGP and PoF3D. This superior pose distribution contributes to our ability to better recover the full object geometry.

\subsection{Ablation Study}
\boldparagraph{Template Feature Field} We now ablate the design choice of using template feature fields for pose estimation.
Alternatively, one may directly obtain a \textit{template radiance field} (TeRF) for pose estimation, without introducing the additional feature branch. This template radiance field and the real RGB images can be leveraged to solve the camera pose estimation task using the method introduced in \cref{sec:pose}. To further enhance the robustness against color variation, one may further convert the RGB images to gray for pose estimation.  
We compare our semantic feature-based method to these two alternatives on the Shapenet Cars dataset with ground truth pose distribution. As shown in \cref{tab:ablation}, directly leveraging a template radiance field for pose estimation leads to worse pose estimation, while our \method is semantically aligned across instances of different appearances and achieves better pose estimation.

\boldparagraph{Freedom of Degrees}
To demonstrate the effect of leveraging four degrees of freedom~(DoF), we conduct experiments on the CompCars dataset since this dataset has a large 2D scale variation. We compare with an alternative that omits the phase correlation for estimating in-plane rotation $\gamma$ and scale $r$. Although training with only two DoF ($\theta, \phi$) achieves lower FID$_{est}$ as shown in \cref{tab:pc_ablation}, we observe that the generated 3D samples have wrong geometry. This observation is also supported by the worse FID$_{360}$ and depth metrics, demonstrating the importance of modeling the scale and in-plane rotation in the camera model.  

\begin{table}[t]
  \centering
  \resizebox{\textwidth}{!}{
  \begin{minipage}[t]{0.44\linewidth}
    \centering
    \resizebox{\textwidth}{!}{
    \begin{tabular}{c|ccc}
      \hline
      Method & TeRF$_{RGB}$ & TeRF$_{Gray}$ & $\method$(Ours)  \\
      $\theta$ KL~$\downarrow$ & 0.0663 & 0.0656 & \textbf{0.0555} \\
      $\phi$ KL ~$\downarrow$ & 0.1422 & 0.1490 & \textbf{0.0696} \\
      \hline
    \end{tabular}
    }
    \captionsetup{font={scriptsize,stretch=1.25},justification=raggedright}
    \caption{Ablation Study on Shapenet Car.}
    \label{tab:ablation}
  \end{minipage}
  \begin{minipage}[t]{0.54\linewidth}
    \centering
    \resizebox{\textwidth}{!}{
    \begin{tabular}{c|cccc}
      \hline
      DoF & Depth$_{360}$~$\downarrow$ & Depth$_{est}$~$\downarrow$ & FID$_{360}$ ~$\downarrow$ & FID$_{est}$ ~$\downarrow$  \\
      $\theta, \phi$ & 4.98 & 1.16 & 39.66 & \textbf{11.09} \\
      $\theta, \phi, \gamma, r$  & \textbf{0.31} & \textbf{0.29} & \textbf{27.31} & 20.60   \\
      \hline
    \end{tabular}
    }
    \captionsetup{font={scriptsize,stretch=1.25},justification=raggedright}
    \caption{Ablation Study on CompCars.}
    \label{tab:pc_ablation}
  \end{minipage}}
  
\end{table}

\section{Limitation}\label{sec:limitation}

\begin{wrapfigure}[7]{r}{0.4\textwidth}
  \centering
  \includegraphics[width=0.4\textwidth]{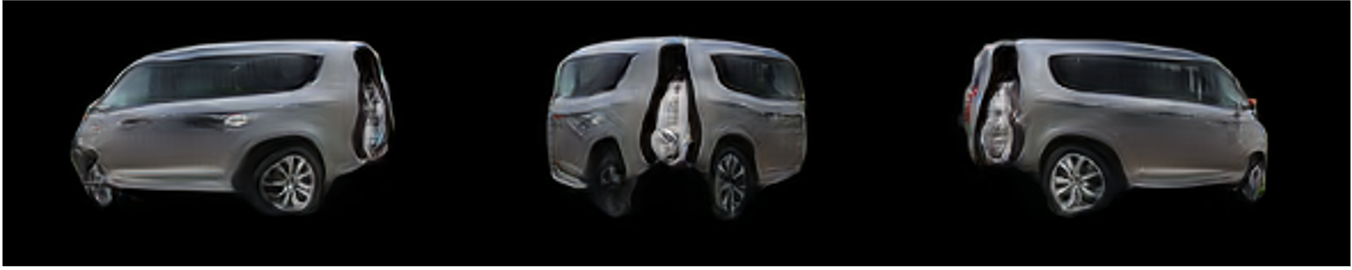}
  \caption{\textbf{Failure case.} }
   \label{fig:failure}
\end{wrapfigure}

Our method can not model images with significant perspective distortion. To fit the distribution of real images, the perspective effect is modeled by generated geometry and leads to the failure case like \figref{fig:failure}. The 2D-3D pose estimation may be influenced by the geometry of the instance since we use the mse to select the best-matching camera pose. In the future, we will explore disentangling the geometry information during the matching process. Besides, we do not model the articulation of the instance. The generator may model different articulations in different views and occasionally lead to instances with multiple legs.

\section{Conclusion}\label{sec:conclusion}

We present \method, a novel approach to enable learning 3D-aware generative models from in-the-wild images with unknown camera pose distribution. By estimating the camera poses of real images on the fly, we demonstrate that our method is capable of recovering the full 3D geometry from datasets of challenging distributions. Despite achieving promising results on in-the-wild images, our method learns a single feature template and hence is only applicable to a specific category. Future work may explore the usage of multiple templates for learning a single 3D-aware generative model on images of multiple categories.
\subsection*{Acknowledgements}
This work is supported by NSFC under grant 62202418 and U21B2004, and partially by Ant Group.

\bibliographystyle{splncs04}
\bibliography{ref.bib}

\begin{thebibliography}{10}
\providecommand{\url}[1]{\texttt{#1}}
\providecommand{\urlprefix}{URL }
\providecommand{\doi}[1]{https://doi.org/#1}

\bibitem{amir2021deep}
Amir, S., Gandelsman, Y., Bagon, S., Dekel, T.: Deep vit features as dense visual descriptors. ECCVW What is Motion For?  (2022)

\bibitem{caron2021emerging}
Caron, M., Touvron, H., Misra, I., J\'egou, H., Mairal, J., Bojanowski, P., Joulin, A.: Emerging properties in self-supervised vision transformers. In: Proceedings of the International Conference on Computer Vision (ICCV) (2021)

\bibitem{EG3D}
Chan, E.R., Lin, C.Z., Chan, M.A., Nagano, K., Pan, B., De~Mello, S., Gallo, O., Guibas, L.J., Tremblay, J., Khamis, S., Karras, T., Wetzstein, G.: Efficient geometry-aware 3d generative adversarial networks. In: CVPR (2022)

\bibitem{Chan2021ARXIV}
Chan, E.R., Lin, C.Z., Chan, M.A., Nagano, K., Pan, B., Mello, S.D., Gallo, O., Guibas, L., Tremblay, J., Khamis, S., Karras, T., Wetzstein, G.: Efficient geometry-aware {3D} generative adversarial networks. In: IEEE Conf. Comput. Vis. Pattern Recog. (2022)

\bibitem{Chan2020CVPR}
Chan, E.R., Monteiro, M., Kellnhofer, P., Wu, J., Wetzstein, G.: Pi-gan: Periodic implicit generative adversarial networks for 3d-aware image synthesis. In: IEEE Conf. Comput. Vis. Pattern Recog. (2021)

\bibitem{Chang2015ARXIV}
Chang, A.X., Funkhouser, T.A., Guibas, L.J., Hanrahan, P., Huang, Q., Li, Z., Savarese, S., Savva, M., Song, S., Su, H., Xiao, J., Yi, L., Yu, F.: Shapenet: An information-rich 3d model repository. arXiv.org  \textbf{1512.03012} (2015)

\bibitem{mmdetection}
Chen, K., Wang, J., Pang, J., Cao, Y., Xiong, Y., Li, X., Sun, S., Feng, W., Liu, Z., Xu, J., Zhang, Z., Cheng, D., Zhu, C., Cheng, T., Zhao, Q., Li, B., Lu, X., Zhu, R., Wu, Y., Dai, J., Wang, J., Shi, J., Ouyang, W., Loy, C.C., Lin, D.: {MMDetection}: Open mmlab detection toolbox and benchmark. arXiv preprint arXiv:1906.07155  (2019)

\bibitem{bib:mimic3d}
Chen, X., Deng, Y., Wang, B.: Mimic3d: Thriving 3d-aware gans via 3d-to-2d imitation. In: Proceedings of the IEEE/CVF International Conference on Computer Vision (ICCV) (2023)

\bibitem{Chen_2023_ICCV}
Chen, X., Huang, J., Bin, Y., Yu, L., Liao, Y.: Veri3d: Generative vertex-based radiance fields for 3d controllable human image synthesis. In: Proceedings of the IEEE/CVF International Conference on Computer Vision (ICCV). pp. 8986--8997 (October 2023)

\bibitem{chen2020deep}
Chen, Z., Xu, X., Wang, Y., Xiong, R.: Deep phase correlation for end-to-end heterogeneous sensor measurements matching. In: Proceedings of the 2020 Conference on Robot Learning. Proceedings of Machine Learning Research, vol.~155, pp. 2359--2375. PMLR (16--18 Nov 2021), \url{https://proceedings.mlr.press/v155/chen21g.html}

\bibitem{DENG2021ARXIV}
Deng, K., Liu, A., Zhu, J., Ramanan, D.: Depth-supervised nerf: Fewer views and faster training for free. arXiv.org  \textbf{2107.02791} (2021)

\bibitem{deng2022gram}
Deng, Y., Yang, J., Xiang, J., Tong, X.: Gram: Generative radiance manifolds for 3d-aware image generation. In: Proceedings of the IEEE/CVF Conference on Computer Vision and Pattern Recognition. pp. 10673--10683 (2022)

\bibitem{Gu2021ARXIV}
Gu, J., Liu, L., Wang, P., Theobalt, C.: Stylenerf: {A} style-based 3d-aware generator for high-resolution image synthesis. Int. Conf. Learn. Represent.  (2022)

\bibitem{Henzler2019ICCV}
Henzler, P., Mitra, N.J., Ritschel, T.: Escaping plato's cave: 3d shape from adversarial rendering. In: Int. Conf. Comput. Vis. (2019)

\bibitem{Heusel2017NIPS}
Heusel, M., Ramsauer, H., Unterthiner, T., Nessler, B., Hochreiter, S.: Gans trained by a two time-scale update rule converge to a local nash equilibrium. In: Adv. Neural Inform. Process. Syst. (2017)

\bibitem{jo20233d}
Jo, K., Jin, W., Choo, J., Lee, H., Cho, S.: 3d-aware generative model for improved side-view image synthesis. In: Proceedings of the IEEE/CVF International Conference on Computer Vision. pp. 22862--22872 (2023)

\bibitem{Jo2021ARXIV}
Jo, K., Shim, G., Jung, S., Yang, S., Choo, J.: Cg-nerf: Conditional generative neural radiance fields. arXiv.org  (2021)

\bibitem{Karras2019stylegan2}
Karras, T., Laine, S., Aittala, M., Hellsten, J., Lehtinen, J., Aila, T.: Analyzing and improving the image quality of {StyleGAN}. In: Proc. CVPR (2020)

\bibitem{lerf2023}
Kerr, J., Kim, C.M., Goldberg, K., Kanazawa, A., Tancik, M.: Lerf: Language embedded radiance fields. In: International Conference on Computer Vision (ICCV) (2023)

\bibitem{kobayashi2022distilledfeaturefields}
Kobayashi, S., Matsumoto, E., Sitzmann, V.: Decomposing nerf for editing via feature field distillation. In: Advances in Neural Information Processing Systems. vol.~35 (2022), \url{https://arxiv.org/pdf/2205.15585.pdf}

\bibitem{Kuglin1975ThePC}
Kuglin, C.D.: The phase correlation image alignment method (1975), \url{https://api.semanticscholar.org/CorpusID:61133413}

\bibitem{Liao2020CVPR}
Liao, Y., Schwarz, K., Mescheder, L., Geiger, A.: Towards unsupervised learning of generative models for 3d controllable image synthesis. In: IEEE Conf. Comput. Vis. Pattern Recog. (2020)

\bibitem{Mescheder2018ICML}
Mescheder, L., Geiger, A., Nowozin, S.: Which training methods for gans do actually converge? In: Int. Conf. Mach. Learn. (2018)

\bibitem{niemeyer2021giraffe}
Michael, N., Andreas, G.: Giraffe: Representing scenes as compositional generative neural feature fields. In: {IEEE} Conference on Computer Vision and Pattern Recognition, {CVPR} 2021. pp. 11453--11464. {IEEE} Computer Society (2021)

\bibitem{Mildenhall2020ECCV}
Mildenhall, B., Srinivasan, P.P., Tancik, M., Barron, J.T., Ramamoorthi, R., Ng, R.: {NeRF}: Representing scenes as neural radiance fields for view synthesis. In: Eur. Conf. Comput. Vis. (2020)

\bibitem{mokady2022selfdistilled}
Mokady, R., Yarom, M., Tov, O., Lang, O., Daniel Cohen-Or, Tali~Dekel, M.I., Mosseri, I.: Self-distilled stylegan: Towards generation from internet photos (2022)

\bibitem{Nguyen-Phuoc2019ICCV}
Nguyen{-}Phuoc, T., Li, C., Theis, L., Richardt, C., Yang, Y.: Hologan: Unsupervised learning of 3d representations from natural images. In: Int. Conf. Comput. Vis. (2019)

\bibitem{Niemeyer2021THREEDV}
Niemeyer, M., Geiger, A.: Campari: Camera-aware decomposed generative neural radiance fields (2021)

\bibitem{OrEl2021ARXIV}
Or-El, R., Luo, X., Shan, M., Shechtman, E., Park, J., Kemelmacher, I.: Stylesdf: High-resolution 3d-consistent image and geometry generation. In: IEEE Conf. Comput. Vis. Pattern Recog. (2022)

\bibitem{Pan2021NEURIPS}
Pan, X., Xu, X., Loy, C.C., Theobalt, C., Dai, B.: A shading-guided generative implicit model for shape-accurate 3d-aware image synthesis. In: Adv. Neural Inform. Process. Syst. (2021)

\bibitem{Peng2018PVNetPV}
Peng, S., Liu, Y., Huang, Q.X., Bao, H., Zhou, X.: Pvnet: Pixel-wise voting network for 6dof pose estimation. 2019 IEEE/CVF Conference on Computer Vision and Pattern Recognition (CVPR) pp. 4556--4565 (2018), \url{https://api.semanticscholar.org/CorpusID:57189382}

\bibitem{Schwarz2020NeurIPS}
Schwarz, K., Liao, Y., Niemeyer, M., Geiger, A.: Graf: Generative radiance fields for 3d-aware image synthesis. In: Adv. Neural Inform. Process. Syst. (2020)

\bibitem{Schwarz2022NEURIPS}
Schwarz, K., Sauer, A., Niemeyer, M., Liao, Y., Geiger, A.: Voxgraf: Fast 3d-aware image synthesis with sparse voxel grids. Adv. Neural Inform. Process. Syst.  (2022)

\bibitem{shi2023pof3d}
Shi, Z., Shen, Y., Xu, Y., Peng, S., Liao, Y., Guo, S., Chen, Q., Yeung, D.Y.: Learning 3d-aware image synthesis with unknown pose distribution  (2023)

\bibitem{shin2023ballgan}
Shin, M., Seo, Y., Bae, J., Choi, Y.S., Kim, H., Byun, H., Uh, Y.: Ballgan: 3d-aware image synthesis with a spherical background. arXiv preprint arXiv:2301.09091  (2023)

\bibitem{3dgp}
Skorokhodov, I., Siarohin, A., Xu, Y., Ren, J., Lee, H.Y., Wonka, P., Tulyakov, S.: 3d generation on imagenet. In: International Conference on Learning Representations (2023), \url{https://openreview.net/forum?id=U2WjB9xxZ9q}

\bibitem{sun2022fenerf}
Sun, J., Wang, X., Zhang, Y., Li, X., Zhang, Q., Liu, Y., Wang, J.: Fenerf: Face editing in neural radiance fields. In: Proceedings of the IEEE/CVF Conference on Computer Vision and Pattern Recognition. pp. 7672--7682 (2022)

\bibitem{tschernezki22neural}
Tschernezki, V., Laina, I., Larlus, D., Vedaldi, A.: Neural feature fusion fields: {3D} distillation of self-supervised {2D} image representations. In: Proceedings of the International Conference on {3D} Vision (3DV) (2022)

\bibitem{Wu2019GH}
Wu, Y., Kirillov, A., Massa, F., Lo, W.Y., Girshick, R.: Detectron2. \url{https://github.com/facebookresearch/detectron2} (2019)

\bibitem{xiang2023gram}
Xiang, J., Yang, J., Deng, Y., Tong, X.: Gram-hd: 3d-consistent image generation at high resolution with generative radiance manifolds. In: Proceedings of the IEEE/CVF International Conference on Computer Vision. pp. 2195--2205 (2023)

\bibitem{Xu2021NEURIPS}
Xu, X., Pan, X., Lin, D., Dai, B.: Generative occupancy fields for 3d surface-aware image synthesis. In: Adv. Neural Inform. Process. Syst. (2021)

\bibitem{Xu2021ARXIV}
Xu, Y., Peng, S., Yang, C., Shen, Y., Zhou, B.: 3d-aware image synthesis via learning structural and textural representations. IEEE Conf. Comput. Vis. Pattern Recog.  (2022)

\bibitem{xue2022giraffe}
Xue, Y., Li, Y., Singh, K.K., Lee, Y.J.: Giraffe hd: A high-resolution 3d-aware generative model. In: Proceedings of the IEEE/CVF Conference on Computer Vision and Pattern Recognition. pp. 18440--18449 (2022)

\bibitem{yang2023emernerf}
Yang, J., Ivanovic, B., Litany, O., Weng, X., Kim, S.W., Li, B., Che, T., Xu, D., Fidler, S., Pavone, M., Wang, Y.: Emernerf: Emergent spatial-temporal scene decomposition via self-supervision. arXiv preprint arXiv:2311.02077  (2023)

\bibitem{CompCars2015CVPR}
Yang, L., Luo, P., Loy, C.C., Tang, X.: A large-scale car dataset for fine-grained categorization and verification. In: {IEEE} Conference on Computer Vision and Pattern Recognition, {CVPR} 2015, Boston, MA, USA, June 7-12, 2015. pp. 3973--3981. {IEEE} Computer Society (2015). \doi{10.1109/CVPR.2015.7299023}, \url{https://doi.org/10.1109/CVPR.2015.7299023}

\bibitem{Wei2021CVPR}
Yin, W., Zhang, J., Wang, O., Niklaus, S., Mai, L., Chen, S., Shen, C.: Learning to recover 3d scene shape from a single image. In: Proc. IEEE Conf. Comp. Vis. Patt. Recogn. (CVPR) (2021)

\bibitem{GET3D--}
Yu, F., Wang, X., Li, Z., Cao, Y., Shan, Y., Dong, C.: {GET3D-:} learning {GET3D} from unconstrained image collections. CoRR  \textbf{abs/2307.14918} (2023). \doi{10.48550/ARXIV.2307.14918}, \url{https://doi.org/10.48550/arXiv.2307.14918}

\bibitem{Yu2015ARXIV}
Yu, F., Seff, A., Zhang, Y., Song, S., Funkhouser, T., Xiao, J.: Lsun: Construction of a large-scale image dataset using deep learning with humans in the loop. arXiv.org  \textbf{1506.03365} (2015)

\bibitem{zhao2022gmpi}
Zhao, X., Ma, F., Güera, D., Ren, Z., Schwing, A.G., Colburn, A.: Generative multiplane images: Making a 2d gan 3d-aware. In: Proc. ECCV (2022)

\bibitem{zhao2022tex}
Zhao, X., Zhao, Z., Schwing, A.G.: Initialization and alignment for adversarial texture optimization. In: European Conference on Computer Vision (ECCV) (2022)

\bibitem{Zhou2021ARXIV}
Zhou, P., Xie, L., Ni, B., Tian, Q.: {{CIPS}}-{{3D}}: A {{3D}}-{{Aware Generator}} of {{GANs Based}} on {{Conditionally}}-{{Independent Pixel Synthesis}}. arXiv.org  (2021)

\end{thebibliography}

\clearpage
\section*{\Large Appendix}
\appendix
\section{Implementation Details}
We evaluate on four datasets, including Shapenet Cars~\cite{Chang2015ARXIV,EG3D}, CompCars~\cite{CompCars2015CVPR}, SDIP Elephant~\cite{mokady2022selfdistilled}, and LSUN Plane~\cite{Yu2015ARXIV}. CompCars contains 136k unposed images capturing the entire cars with different styles. The original dataset contains images with different aspect ratios. We preprocess the images by center cropping, padding to the squared images with the same length, and resizing them to 256$\times$256. We use the mask to set the background black and filter the data with bad mask estimation and extreme scale, leading to 110k images. 
LSUN Plane is a dataset that contains unposed images of different planes. We use MMDetection~\cite{mmdetection} to detect the plane and filter the plane larger than 226$\times$226 resolution and the occluded plane, leading to 130k images. We rescale the plane to make the large side equal to 226 and padding it to 256 resolution. 

\section{Runtime Analysis}
\boldparagraph{Runtime Breakdown} We report the average runtime of different processes during training in \cref{tab:time}. The runtime analysis is conducted on an A6000 GPU. The "Template Rendering" indicates the time to render the template feature field at discretized azimuth and elevation angles $\theta$ and $\phi$ to obtain 2D template features $\{\overline{\bF}\}_{k=1}^{N_{\theta}\times N_{\phi}}$. The batch size to render the template is 32. We update the template and render it once every 16 iterations before 3k iterations and then once every epoch. The template rendering time in the early stage is averaged over 16 iterations. Since the iterations of each epoch are different for different datasets, we report the averaged template rendering time in the late stage using the dataset of the smallest amount of images, \ie SDIP Elephant dataset. The "Phase Correlation" refers to the time for estimating the scale $r$ and in-plane rotation $\gamma$, and warping each feature template to yield $\{\tilde{\bF}\}_{k=1}^{N_{\theta}\times N_{\phi}}$. The "Camera Pose Sampling" includes the time to calculate mean square error and to perform inverse sampling (see Eq. 2 and Eq. 3 of the main paper). "Training" indicates the training time without camera pose estimation, which includes the data loading time, the network forward time, and the optimization time. The batch size for training is 4. Note that we use a fixed camera pose and do not perform pose estimation after 500k iterations. Compared to the overall iterations of 6250k, the increased time is acceptable. 

\boldparagraph{Runtime Comparison with Na\"ive Grid Search} As mentioned in our main paper, one can implement a na\"ive grid search method by discretizing all four variables $(\theta, \phi, \gamma, r)$ we consider for the camera poses. We demonstrate that this na\"ive approach significantly increases the pose estimation time in \cref{tab:naive time}. Here, we further discretize scale $r$ and in-plane rotation $\gamma$, each discretized into $256$ values, increasing the total amount of 2D feature templates by $256^2$ times. While this omits the phase correlation module, it significantly increases the template rendering time and is hence intractable for training the 3D GAN.

\begin{table}[h]
    \centering
    \resizebox{\textwidth}{!}{
    \begin{tabular}{c|c|c|c|c|c}
      \hline
       Process & \makecell[c]{Template Rendering \\Early Stage} & \makecell[c]{Template Rendering \\Late Stage}  & \makecell[c]{Phase \\Correlation } & \makecell[c]{Camera Pose\\ Sampling} & Training  \\
      Time (s/iter) & 0.0992 & 0.0023 & 0.3898 & 0.0156 & 1.4038 \\

      \hline
    \end{tabular}
    }
    \caption{\textbf{Time Analysis of Different Processes.}}
    \label{tab:time}
\end{table}

\begin{table}[h]
    \centering
    \resizebox{\textwidth}{!}{
    \begin{tabular}{c|c|c|c|c|c}
      \hline
       Process & \makecell[c]{Template Rendering \\Early Stage} & \makecell[c]{Template Rendering \\Late Stage}  &  \makecell[c]{Phase \\Correlation } & \makecell[c]{Camera Pose\\ Sampling} & Training  \\
      Time (s/iter) & 4423.0519 & 102.5506 &  -- & 74.8994 & 1.4038 \\

      \hline
    \end{tabular}
    }
    \caption{\textbf{Time Analysis of Na\"ive Grid Search.}}
    \label{tab:naive time}
\end{table}

\section{Ablation Study of the Number of  Discretizations}
We conduct an ablation study on the number of discretizations on ShapeNet Cars. We report FID and early-stage template rendering time in \cref{tab:discretization_number}. Reducing discretized bins of $\theta \times \phi$ from $36\times18$ to $24\times12$ or lower worsens FID due to larger quantization steps. Increasing bins beyond $36\times18$ obtains comparable results but yields increasing costs. Note that our method with $12\times6$ bins still outperforms PoF3D in terms of FID$_{gt}$.
\begin{table}[h]
  \centering
  \begin{tabular}{c|c|c|c|c|c}
    \hline
    $\theta \times \phi$ & 12$\times$6 & 24$\times$12 & 36$\times$18 & 48$\times$24 & 60$\times$30 \\
    \hline
     FID$_{gt}$ / FID$_{est}$ & 11.29 / 9.57 & 7.48 / 7.35 & 5.95 / 6.55 & 5.51 / 6.25 & 5.96 / 6.48 \\
    \hline
    time s/iter & 0.0559 & 0.2126 & 0.4966 & 0.8465 & 1.3205 \\
    \hline
  \end{tabular}
  \caption{\small{Different number of bins for discretization.}}
  \label{tab:discretization_number}
\end{table}

\section{Comparison on the FFHQ and AFHQ datasets} 
We evaluate our method on the FFHQ and AFHQ datasets. We set the azimuth$\theta$ range as 120 degrees according to the distribution prior for the datasets, and discretize it into 36 values. Since the elevation variation is small, we directly set the elevation angle as 90 degrees.

\begin{figure}[h]
  \centering
  \includegraphics[width=1.0\linewidth]{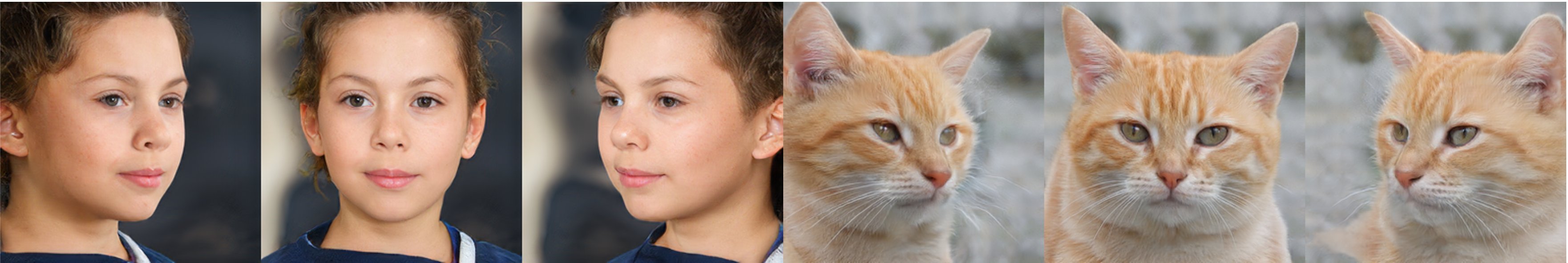}
   \caption{Our generated samples on face and cat datasets.}
   \label{fig:face_cat}
\end{figure}

We achieve comparable results with PoF3D on the FFHQ dataset and better results on the AFHQ dataset, as shown in \cref{tab:face_cat}. Despite PoF3D performing well in generated poses, their results degenerate in GT poses. We achieve comparable results with EG3D on the AFHQ dataset, even though EG3D uses GT poses (with pose condition) and we do not. Despite not using GT poses, our result is only slightly worse than EG3D on the FFHQ dataset. \cref{fig:face_cat} demonstrates our qualitative results.

\begin{table}[h]
  \centering
  \begin{tabular}{c|cccc|cc}
    \hline
     & \multicolumn{4}{c|}{FFHQ}& \multicolumn{2}{c}{AFHQ} \\
    Method & Depth$_{gt}$$\downarrow$ & Depth$_{est}$ $\downarrow$ & FID$_{gt}$ $\downarrow$ &  FID$_{est}$ $\downarrow$
     & FID$_{gt}$$\downarrow$ &  FID$_{est}$ $\downarrow$\\
    \hline
    EG3D  & \textbf{0.29} & - & \textbf{4.80} & - & 5.56 & - \\
    PoF3D & 0.37 & \textbf{0.29} & 5.13 & \textbf{4.99} & 16.95 & 5.46  \\ 
    Ours  & 0.36 & 0.35 & 5.64 & 5.37 & \textbf{4.52} & \textbf{4.37}  \\ 
    \hline
  \end{tabular}
  \caption{\small{Comparison on the FFHQ and the AFHQ datasets.}}
  \label{tab:face_cat}
\end{table}

\section{Qualitative Results of Camera Pose Estimation}
\begin{figure}[t]
    \centering
     \def\mywidth{8.9cm}
     \begin{tabular}{P{0.5cm}P{\mywidth}}
     \rotatebox{90}{Real} &\includegraphics[width=\mywidth]{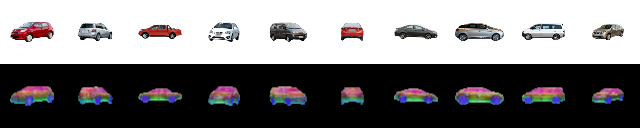} \\
    \rotatebox{90}{Pose}
    &\includegraphics[width=\mywidth]{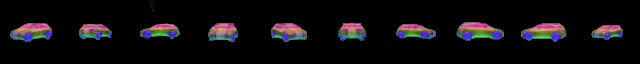} \\
      \rotatebox{90}{Real} &\includegraphics[width=\mywidth]{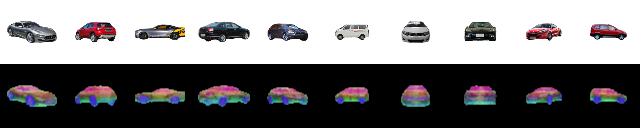} \\
    \rotatebox{90}{Pose}
    &\includegraphics[width=\mywidth]{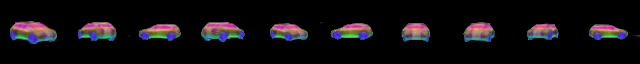} \\
      \rotatebox{90}{Real} &\includegraphics[width=\mywidth]{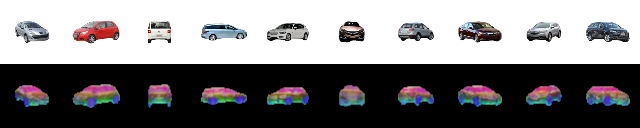} \\
    \rotatebox{90}{Pose}
    &\includegraphics[width=\mywidth]{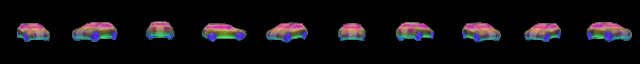} \\

     \end{tabular}
     \caption{\textbf{Qualitative Pose Estimation on CompCars.} }
   \label{fig:pose_CompCars}
    \end{figure}

\begin{figure}[h]
    \centering
     \def\mywidth{8.9cm}
     \begin{tabular}{P{0.5cm}P{\mywidth}}
     \rotatebox{90}{Real} &\includegraphics[width=\mywidth]{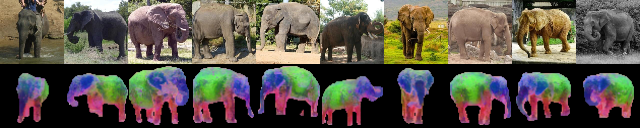} \\
    \rotatebox{90}{Pose}
    &\includegraphics[width=\mywidth]{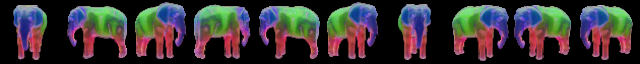} \\
      \rotatebox{90}{Real} &\includegraphics[width=\mywidth]{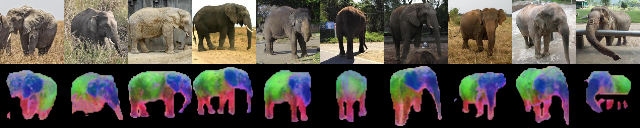} \\
    \rotatebox{90}{Pose}
    &\includegraphics[width=\mywidth]{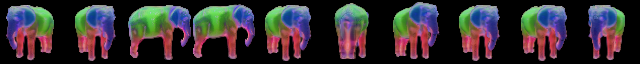} \\
      \rotatebox{90}{Real} &\includegraphics[width=\mywidth]{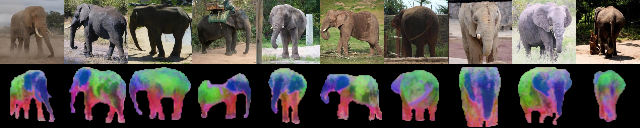} \\
    \rotatebox{90}{Pose}
    &\includegraphics[width=\mywidth]{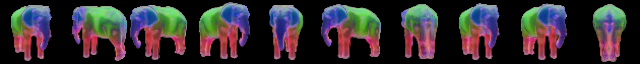} \\

     \end{tabular}
     \caption{\textbf{Qualitative Camera Pose Estimation on SDIP Elephant.} }
   \label{fig:pose_elephant}
    \end{figure}

\begin{figure}[h]
    \centering
     \def\mywidth{8.9cm}
     \begin{tabular}{P{0.5cm}P{\mywidth}}
     \rotatebox{90}{Real} &\includegraphics[width=\mywidth]{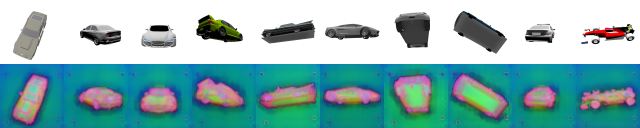} \\
    \rotatebox{90}{Pose}
    &\includegraphics[width=\mywidth]{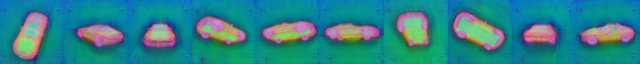} \\
      \rotatebox{90}{Real} &\includegraphics[width=\mywidth]{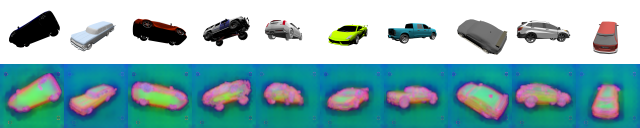} \\
    \rotatebox{90}{Pose}
    &\includegraphics[width=\mywidth]{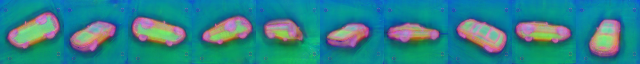} \\
      \rotatebox{90}{Real} &\includegraphics[width=\mywidth]{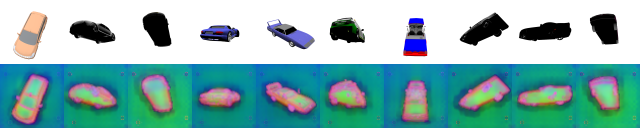} \\
    \rotatebox{90}{Pose}
    &\includegraphics[width=\mywidth]{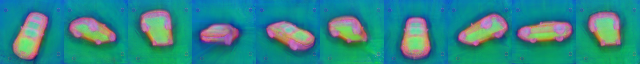} \\

     \end{tabular}
     \caption{\textbf{Qualitative Camera Pose Estimation on Shapenet Cars.} }
   \label{fig:pose_Shapenet}
    \end{figure}

\begin{figure}[h]
    \centering
     \def\mywidth{8.9cm}
     \begin{tabular}{P{0.5cm}P{\mywidth}}
     \rotatebox{90}{Real} &\includegraphics[width=\mywidth]{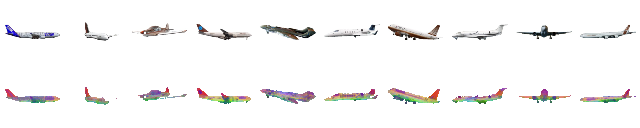} \\
    \rotatebox{90}{Pose}
    &\includegraphics[width=\mywidth]{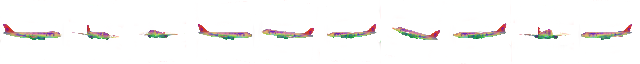} \\
      \rotatebox{90}{Real} &\includegraphics[width=\mywidth]{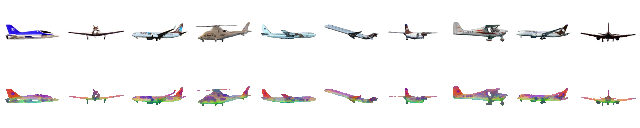} \\
    \rotatebox{90}{Pose}
    &\includegraphics[width=\mywidth]{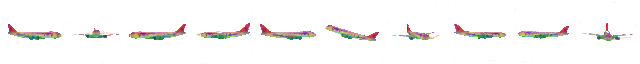} \\
      \rotatebox{90}{Real} &\includegraphics[width=\mywidth]{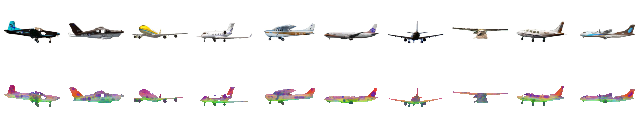} \\
    \rotatebox{90}{Pose}
    &\includegraphics[width=\mywidth]{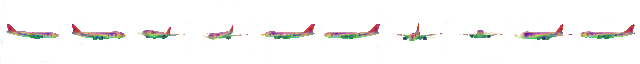} \\

     \end{tabular}
     \caption{\textbf{Qualitative Camera Pose Estimation on LSUN Plane.} }
   \label{fig:pose_plane}
    \end{figure}

\begin{figure}[h]
    \centering
     \def\mywidth{.24}
     \begin{tabular}{cccc}
     \includegraphics[width=\mywidth\linewidth]{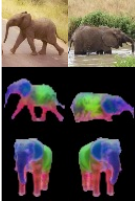} &
     \includegraphics[width=\mywidth\linewidth]{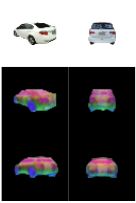} &
      \includegraphics[width=\mywidth\linewidth]{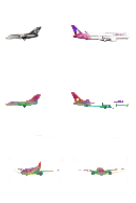} &
       \includegraphics[width=\mywidth\linewidth]{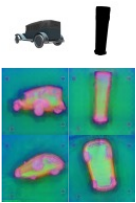}\\
       \begin{small}Elephant\end{small} &
      \begin{small}CompCars\end{small} &
      
      \begin{small}Plane\end{small} &
      \begin{small}ShapenetCars\end{small}
      
     \end{tabular}
     \caption{\textbf{Failure Camera Pose Estimation}}
   \label{fig:failure_pose}
    \end{figure}

\figref{fig:pose_CompCars}, \figref{fig:pose_elephant}, \figref{fig:pose_Shapenet} and \figref{fig:pose_plane} show the estimated poses of real images on CompCars, SDIP Elephant, Shapenet Cars and LSUN Plane, respectively. The first row is the real image $\bI$ and its corresponding DINO feature $\bF$, and the second row is the best-matching template feature $\tilde{\bF}^{\ast}_k$. The corresponding camera pose of $\tilde{\bF}^{\ast}_k$ indicates the estimated pose of the real image. Note that we only perform phase correlation on CompCars and LSUN Plane since Shapenet Cars and SDIP Elephant do not have large variations in scale and in-plane rotation. The results demonstrate that our method can perform fairly accurate camera pose estimation. \figref{fig:failure_pose} shows the failure cases of pose estimation. The estimated pose may not be accurate due to the object articulation, partial observation, significant geometry difference, lack of modeling object translation, etc. This could be solved by introducing multiple templates, modeling more freedom of degrees of camera pose, and disentangling the geometry information during the matching process in the future. We remark that despite some instances of inaccurate pose estimation, the overall distribution of poses is generally precise. This accuracy enables the trained generator to produce objects with complete geometry and the capability for comprehensive 360-degree view synthesis.

\section{Uncurated Qualitative Results}
\begin{figure}[h]
    \centering
     \def\mywidth{12cm}
     \includegraphics[width=\mywidth]{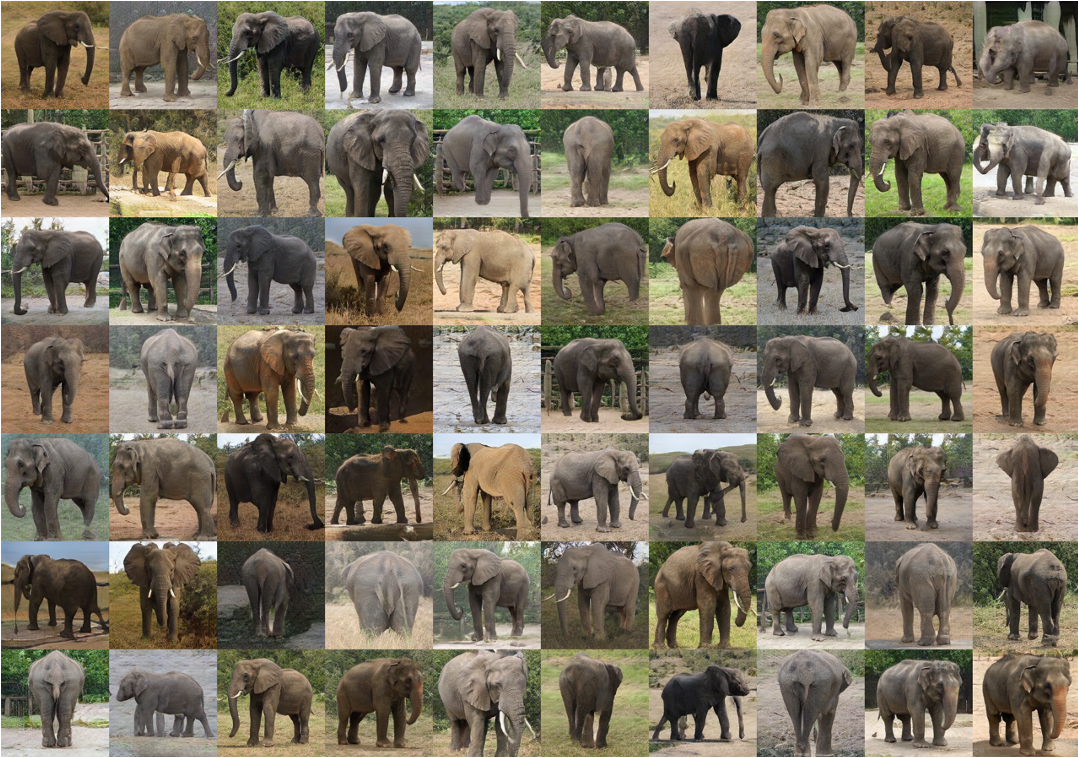} \\
     \caption{\textbf{Uncurated Result on SDIP Elephant.} }
   \label{fig:Uncurated_Elephant}
    \end{figure}

\begin{figure}[h]
    \centering
     \def\mywidth{12cm}
     \includegraphics[width=\mywidth]{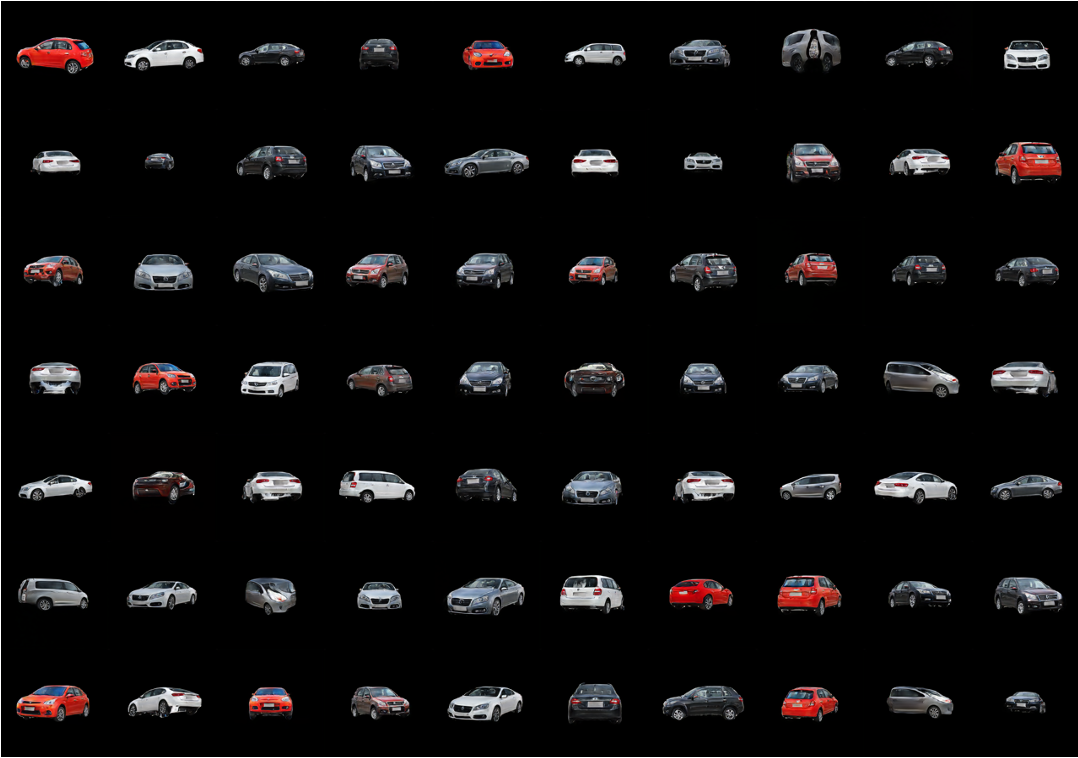} \\
     \caption{\textbf{Uncurated Result on CompCars.} }
   \label{fig:Uncurated_CompCars}
    \end{figure}

\begin{figure}[h]
    \centering
     \def\mywidth{12cm}
     \includegraphics[width=\mywidth]{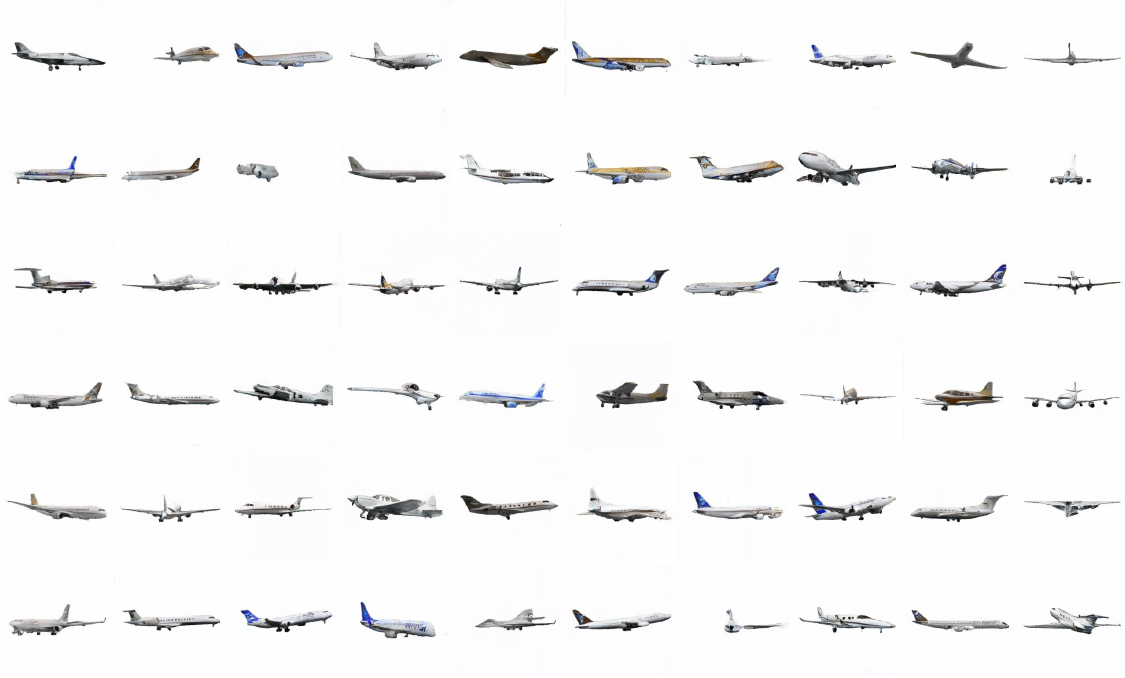} \\
     \caption{\textbf{Uncurated Result on LSUN Plane.} }
   \label{fig:Uncurated_Plane}
    \end{figure}

\begin{figure}[h]
    \centering
     \def\mywidth{12cm}
     \includegraphics[width=\mywidth]{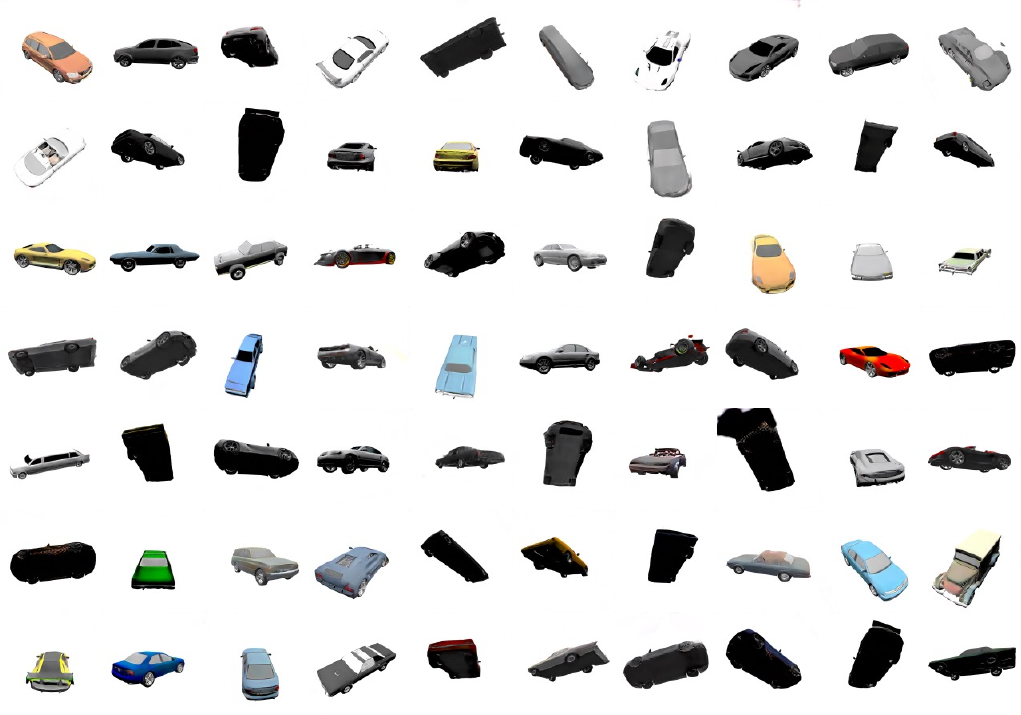} \\
     \caption{\textbf{Uncurated Result on Shapenet Cars.} }
   \label{fig:Uncurated_Shapenet}
    \end{figure}

\figref{fig:Uncurated_Elephant}, \figref{fig:Uncurated_CompCars}, \figref{fig:Uncurated_Plane} and \figref{fig:Uncurated_Shapenet} show the uncurated qualitative results on SDIP Elephant, CompCars, LSUN Plane and Shapenet Cars. Our results demonstrate good fidelity and diversity.

\end{document}